\RequirePackage{fix-cm}
\documentclass{svjour3}                     
\smartqed  
\usepackage{graphicx}
\usepackage{cite}
\usepackage{amssymb,bm}
\usepackage{amsmath}
\usepackage[caption=false,font=normalsize,labelfont=sf,textfont=sf]{subfig}
\DeclareMathOperator*{\argmax}{arg\,max}
\newtheorem{assumption}{Assumption}
\DeclareMathOperator{\sign}{sgn}
\graphicspath{ {figures/} }

\begin{document}
\title{Incorporating Non-Parametric Knowledge to the Least Mean Square Adaptive Filter
}



\title{Robust Non-parametric Knowledge-based Diffusion Least Mean Squares over Adaptive Networks}


\author{Soheila~Ashkezari-Toussi \and Hadi~Sadoghi-Yazdi }


\institute{Soheila~Ashkezari-Toussi \\ \email{sohei.ashkezari@mail.um.ac.ir}\\ \\
	Hadi~Sadoghi-Yazdi\\
	\email {h-sadoghi@um.ac.ir}\\
	\\ \\
	\at{Department of Computer Engineering, Ferdowsi University of Mashhad, Mashhad, Iran} 
	\at{Center of Excellence on Soft Computing and Intelligent Information Processing, Ferdowsi University of Mashhad, Mashhad, Iran}
}

\date{Received: date / Accepted: date}

\maketitle

\begin{abstract}
The present study proposes incorporating of non-parametric knowledge to the diffusion least-mean-squares algorithm in the framework of a maximum a posteriori (MAP) estimation. The proposed algorithm leads to a robust estimation of an unknown parameter vector in a group of cooperative estimators. 
Utilizing kernel density estimation and buffering some intermediate estimations, the prior distribution and conditional likelihood of the parameters vector in each node are calculated. Pseudo Huber loss function is used for designing the likelihood function. Also, an error thresholding function is defined to reduce the computational overhead as well as more relaxation against noise, which stops the update every time an error is less than a predefined threshold.
The performance of the proposed algorithm is examined in the stationary and non-stationary scenarios in the presence of Gaussian and non-Gaussian noise. Results show the robustness of the proposed algorithm in the presence of different noise types.

\end{abstract}






 \section{Introduction}
\textit{Adaptive networks} consist of a set of nodes that are linked together and cooperate with neighboring nodes to respond in real time to the streaming measurements in the stationary and non-stationary environments. 
Distributed processing techniques rely on local cooperation and data processing. Each node estimates the parameters of interest from local observations. All elements can interact with their neighboring nodes according to the network topology. In this way, each node can receive information from its adjacent sensor nodes and this information can be processed to obtain a signal estimator based not only on its own information but also on that from its neighbors. Therefore, in contrast with a centralized parameter estimation, there is not a single fusion center in the network; rather, each node acts as both a sensor and a fusion center  \cite{R1}. The distributed approach significantly reduces the communication and processing overhead and has been widely used in different applications, such as cognitive radios \cite{di2011}, mobile adaptive networks \cite{tu2011,bazzi2015}, environmental monitoring \cite{cao2010,duan2017}, industrial automation \cite{chen2010,bai2018}, decision-making \cite{tu2014,khawatmi2017}, modeling bird flight formations \cite{cattivelli2011}.

\textit{Diffusion strategy} is a well-known approach over distributed adaptive networks \cite{R8,R9,R21,LEE2015,chen2014,chen2015}. Since variables in many real applications are contaminated by \textit {random noise}, processes perform in a non-deterministic way. 
Dealing with noisy measurement is one of the important challenges in the literature of this area. 
To overcome this problem some researches try to minimize the modified  error functions while  another group includes algorithms with cost functions based on information theory learning (ITL). In the  error minimization approach there are extended versions of well-known algorithms such as diffusion LMS (DLMS) \cite{R8} and diffusion recursive least square (DRLS) \cite{cattivelli2008}. Clearly, MSE (mean square error)-based algorithms can not guarantee to converge when the environment is contaminated by non-Gaussian noise.  
In such cases, other power of error are used in some researches like \cite{wen2013,ni2016}.  
In \cite{ni2016b,seo2016diffusion,shi2017two,gao2018steady} sign operator has been used to make DLMS robust against non-Gaussian noise.  In \cite{chen2018} the diffusion least logarithmic absolute difference (DLLAD)
algorithm has been proposed which adopts both the logarithm operation and sign operation to the error.
On the other hand, the main contribution of ITL-based algorithms is using entropy or corentropy in the cost function for dealing with non-Gaussian noise. In \cite{bazzi2015robust,ma2016}  diffusion maximum correntropy criterion (DMCC) algorithm has been proposed to improve the performance of the distributed estimation over network in the impulsive noise environments. The error entropy criterion based on the minimum error entropy (MEE) has been proposed in \cite{li2013} which achieves better results comparing with MSE-based algorithms under non-Gaussian noise. 

\textit{Probabilistic modeling} is an appropriate choice to express the inherent uncertainty exists in data. It can naturally deal with missing values \cite{Ilin}, reduce the computational burden using EM algorithm \cite{Babadi}, is used to extend single model structure to mixture model cases to handle more complicated problems and avoids overfitting taking advantages of Bayesian methods for model selection and parameter tuning \cite{Arenas}. Also, it provides a clear separation of the model and its algorithm \cite{Candy}. 
Utilizing probabilistic models facilitate incorporating the \textit{prior knowledge} in the learning process. A way for incorporation the prior are adjustable weight coefficients that are assigned different learning rates. In \cite{R17}, it is assumed that an unknown parameter vector has a probability density function and the negative logarithmic probability of its distribution is used as prior knowledge. Then, by using a weighted stochastic gradient, this knowledge is utilized in the adaption algorithm. In \cite{R18}, regularization terms have been included as the probability distribution of the filter coefficients. The adaptive filter is designed using Gaussian distribution and applying a maximum a posteriori estimation. Usage of previous estimations as prior information for the next estimation in the Kalman filter is discussed in \cite{R19}.  Probabilistic DLMS proposed in \cite{guan2019} approximates the posterior distribution with an isotropic Gaussian distribution.
 Recently, the \textit{non-parametric probabilistic least mean square (NPLMS) adaptive filter} has been proposed in \cite{ashkezari2018} for the estimation of an unknown parameter vector from noisy measurements. The NPLMS combines parameter space and signal space by combining the prior knowledge of the probability distribution of the process with the evidence existing in
the signal. Taking advantage of kernel density estimation and buffering some of the intermediate estimations the prior distribution has been estimated. ‌Benefiting the probabilistic modeling, the NPLMS is robust against the Gaussian and non-Gaussian noise. This paper extends the NPLMS algorithm over distributed adaptive networks. Therefore, the non-parametric probabilistic diffusion least mean square (NPDLMS) has been proposed to overcome noisy environments over adaptive networks. Furthermore, to design the likelihood function it has been proposed to use a seudo-Huber loss function \cite{hartley2003} which is robust to deal with different type of noise in stationary and non-stationary environments. Utilizing pseudo-Huber loss function  in Diffusion LMS has been investigated in \cite{ashkezari2019} to create a robust algorithm against noise in adaptive networks. Therefore, the main contribution of this manuscript is to extend the NPLMS algorithms \cite{ashkezari2018} over distributed adaptive networks and benefit pseudo-Huber loss function \cite{ashkezari2019} to design the likelihood function.

The rest of this manuscript has been organized as follow. The non-parametric probabilistic diffusion least mean square has been proposed in Section 2. The performance of the proposed algorithm is analyzed in Section 3. 
Section 4 presents how to reduce the computational overhead of the proposed algorithm. Simulation results are presented in Section 5. Finally, Section 6 concludes the paper.

\textit{Notations:} Let ${\rm \mathbb{R}}$ denotes the set of real numbers. Matrices are represented by uppercase fonts and vectors by lowercase fonts. Boldface letters represent random variables and normal letters stand for deterministic variables. The superscript $(\cdot )^{T} $ denotes the transpose of a matrix or a vector. Symbols $Tr\left(\cdot \right)$,$\rho \left(\cdot \right)$and $\lambda _{k} \left(\cdot \right)$represent the trace, spectral radius, and the \textit{k}th eigenvalue of their matrix argument respectively. The expectation of a matrix is represented by $\mathbb{E}\left[\cdot \right]$. The Kronecker product is denoted by$\otimes $.The operator $diag\left\{\cdot \right\}$ converts its arguments into a block diagonal matrix and $vec\left(\cdot \right)$is the vectorization of matrices. $col\left\{\cdot \right\}$ is a vector obtained by stacking the specified vectors. If $\Sigma $is a matrix, $\left\| x\right\| _{\Sigma }^{2} =x^{T} \Sigma x$ is utilized for the weighted square norm of$x$. If $\sigma $ is a vector, the notation $\left\| x\right\| _{\sigma } $ represents$\left\| x\right\| _{diag\left\{\sigma \right\}} $.

\section{Method}\label{}
\subsection{ System model and assumptions}
Here, $N$ local filters are assumed for computing the unknown parameter vector $\boldsymbol \theta_o $. Consider a network consisting of $N$connected estimators which are labeled as $S_{k} , k=1,...,N$. A network is presented by graph $G$ which consists of $N$ vertices (representing the estimators) and a set of edges connecting the elements to each other. A self-loop shows an edge that connects an estimator to itself.
The neighborhood of element $S_{k} $ is denoted by $N_{k} $ and it consists of all estimators that are connected to $S_{k} $ by an edge, including $S_{k} $ itself. The cardinality of $N_{k} $ is denoted by $\left|N_{k} \right|$. Any two neighboring estimators, $S_{k} $ and $S_{l} $, have the ability to share information over the edges connecting them. 
A pair of nonnegative scaling weights $\left\{a_{kl} ,a_{lk} \right\}$ is assigned to the edge connecting $S_{k} $ and $S_{l} $. It is assumed that scaling weights build an affine combination, i.e. $\sum _{l\in N_{k} }a_{lk} =1 $. As illustrated, scalar $a_{lk} $ determines the weights of estimator $S_{l} $ in interaction with estimator $S_{k} $.
At every time instance $n\in \mathbb{ N}$, each estimator $S_{k} ,k=1,...,N$ receives a vector random process ${\boldsymbol{u}}_{k,n} \in \mathbb{R}^{1\times d} $ and the target value ${\boldsymbol{d}}_{k} (n)\in \mathbb{R}$ in the form of $D_{k} (n)=\left\{{\boldsymbol{u}}_{k,n} ,{\boldsymbol{d}}_{k} (n)\right\}$. The measured data $\left\{{\boldsymbol{u}}_{k,n} ,{\boldsymbol{d}}_{k} (n)\right\}$ are linearly modeled as
\begin{equation} \label{eqModel} 
{\boldsymbol{d}}_{k} (n)={\boldsymbol{u}}_{k,n} \boldsymbol{\theta}_{o} +{\boldsymbol{v}}_{k} (n) 
\end{equation} 

where $k$ is the estimator index $S_{k} $, $n$ is time/sample index, $\boldsymbol{\theta}_{o} \in \mathbb{R}^{d\times 1} $ is an unknown parameter vector to be estimated, and scalar ${\boldsymbol{v}}_{k} (n)\in \mathbb{R}$ is the additive zero mean noise with variance $\sigma ^{2} _{v,k} $.\footnote{Subscripts are used to refer to time indices of vector variables and parentheses to refer to the time indices of scalar variables.}
The following assumptions are considered for the model presented in \eqref{eqModel}:
\begin{assumption} 
	The input regression data vectors are spatially independent and identically distributed (i.i.d.) over time, i.e. $R_{k,l} =\mathbb{E}\left\{{\boldsymbol{u}}_{k,n}^{T} {\boldsymbol{u}}_{l,n} \right\}=0,{\rm \; \; }k\ne l$, where $0$ is a vector/matrix with appropriate dimensions and all elements equal zero. Furthermore, they are zero-mean with a positive-definite covariance matrix $R_{u,k} =\mathbb{E}\left\{{\boldsymbol{u}}_{k,n}^{T} {\boldsymbol{u}}_{k,n} \right\}>0$, and $\mathbb{E}$ is expectation operator.
	\label{ass1}
\end{assumption}
\begin{assumption}
	The output additive noise $v_{k} (n)$ is assumed to be temporally white and spatially independent, i.e. \[\mathbb{E}\left\{{\boldsymbol{v}}_{k} \left(n\right){\boldsymbol{v}}_{l} \left(n'\right)\right\}=\left\{\begin{array}{cc} {0} & {k=l,n\ne n'} \\ {0} & {k\ne l} \end{array}\right. \] 
	\label{ass2}
\end{assumption}
\begin{assumption} 	\label{ass3}
	Random variables $u_{k,n} $ and $v_{k} (n)$ are independent for all $k$ and $l$ i.e. $\mathbb{E}\left\{{\boldsymbol{v}}_{k} \left(n\right){\boldsymbol{u}}_{k,n} \right\}=0, ~~ for ~all~k~and~l$.
\end{assumption}
\subsection{ Non-parametric probabilistic diffusion least mean square}

Consider there is a group of $N$ cooperative estimators which works together to contribute to a shared goal, i.e. the estimation of unknown parameter vector $\boldsymbol{\theta}_{o} $. In this way, estimators interact with each other by passing related information among them and  adapt themselves to incoming data to produce a better description of the measured data. They try to compensate for their error in the estimation process with this interaction.
The global cost function $J\left(\theta \right)^{G} $, is defined as
\begin{equation} \label{eqJglobal} 
J\left(\theta \right)^{G} =\sum _{j=1}^{N}J_{j} (\boldsymbol{\theta})  
\end{equation} 
which $J\left(\theta \right)^{G}$ is approximated by a set of alternative local cost function in which each node optimizes its own function. Therefore, the objective function of the \textit{kth} estimator $S_{k} $ is defined as theorem \ref{objfun}.
\begin{theorem} \label{objfun}
	The local objective function at the \textit{kth} estimator $S_k$ is the maximization of the posterior distribution
	\begin{equation}\label{eqPosterior}
	\mathop{\argmax }\limits_{\boldsymbol{\theta}_k }  f\left(\boldsymbol{\theta}_k |\boldsymbol{\Theta} _{k},D_n\right)\propto \mathop{\argmax }\limits_{\boldsymbol{\theta}_k } f\left(\boldsymbol{\Theta} _{k},D_n|\boldsymbol{\theta}_k\right)f\left(\boldsymbol{\theta}_k\right)
	\end{equation}
	where $\boldsymbol{\Theta} _{k} =\left\{\boldsymbol{\theta}_l |l\in N_{k} \setminus k\right\}$ is the set of all neighbor estimators' parameter vector $\boldsymbol{\theta}_{l} $ except $\boldsymbol{\theta}_k$, and $D_{n} =\bigcup _{l\in N_k}D_{l,n}  $ where $D_{l,n} $ is the given local dataset for estimator $S_{l} $ at \textit{nth} iteration. Equation \eqref{eqPosterior} is approximated as
	\begin{equation} \label{eqLocalObjFun} 
	J_{k} \left(\boldsymbol{\theta}\right)\buildrel\Delta\over=   f\left(\boldsymbol{\theta}_k\right)^{2-|N_k|}f\left(D_{k,n}|\boldsymbol{\theta}_k\right)\prod_{l\in\boldsymbol{N_k}\setminus k}
	f\left(\boldsymbol{\theta}_k|\boldsymbol{\theta}_l\right)
	f\left(D_{l,n}|\boldsymbol{\theta}_k\right)
	f\left(\boldsymbol{\theta}_l\right)
	\end{equation} 

	proof: It is assumed that neighbors estimation $\boldsymbol{\theta}_l\in \boldsymbol{\Theta} _{k} $ and $D_n$ are independent given $\boldsymbol{\theta}_{k}$. Also, it is assumed that neighbors $S_l,l\in N_k$ of $S_k$ are independent given the estimated parameter vector $\boldsymbol{\theta}_{k}$. Therefore,
	\begin{equation}\label{eqJointCondDis}
	f\left(\boldsymbol{\Theta} _{k},D_n|\boldsymbol{\theta}_k\right)=f \left(\boldsymbol{\Theta} _{k}\right|\boldsymbol{\theta}_{k})f\left(D_{n}|\boldsymbol{\theta}_k\right)=
	\prod_{\boldsymbol{\theta}_{l}\in\boldsymbol{\Theta} _{k}}f\left(\boldsymbol{\theta} _{l}|\boldsymbol{\theta}_k\right)\prod_{l\in N_{k}} f\left(D_{l,n}|\boldsymbol{\theta}_k\right)
	\end{equation}
	 Using Bayes theorem,
	\begin{equation}\label{eqThetaCondDis}
	f \left(\boldsymbol{\Theta} _{k}\right|\boldsymbol{\theta}_{k})=\prod_{\boldsymbol{\theta}_{l}\in\boldsymbol{\Theta} _{k}}f\left(\boldsymbol{\theta}_{l}|\boldsymbol{\theta}_{k}\right)=f\left(\boldsymbol{\theta}_{k}\right)^{1-|N_k|} \prod_{\boldsymbol{\theta}_{l}\in\boldsymbol{\Theta} _{k}}
	f\left(\boldsymbol{\theta}_{k}|\boldsymbol{\theta}_{l}\right)f\left(\boldsymbol{\theta}_{l}\right)
	\end{equation}
	Substituting \eqref{eqThetaCondDis} in \eqref{eqJointCondDis}, and its result in \eqref{eqPosterior} gives birth to \eqref{eqLocalObjFun}.
\end{theorem}
\begin{theorem}
	The logarithm of the local objective function \eqref{eqLocalObjFun} is defined as
	\begin{equation}\label{eqLogLocalObjFun}
	\begin{split}
		\hat{J_k}({\boldsymbol{\theta}})&=\log J_k({\boldsymbol{\theta}}) \\
		&= \sum_{l\in N_k}\{\log f(D_l|\boldsymbol{\theta}_k)+\log f(\boldsymbol{\theta}_l)\}\\
		&+\sum_{l\in N_k \setminus k}\{log f(\boldsymbol{\theta}_k|\boldsymbol{\theta}_l)-log f(\boldsymbol{\theta}_k)\}
	\end{split}
	\end{equation}
	proof: Computing the logarithm of \eqref{eqLocalObjFun} is straightforward. 
\end{theorem}

For calculating the distribution functions one has to first define
\begin{equation}\label{eqKernel}
K_t(\boldsymbol x-\boldsymbol y)=\frac{1}{t}exp(-\frac{1}{2t}\Arrowvert \boldsymbol x-\boldsymbol y\Arrowvert^2).
\end{equation}

Utilizing kernel density estimation leads to
\begin{equation}\label{eqF_theta}
f(\boldsymbol{\theta}_k)=\frac{1}{B}\sum_{i=1}^{B}K_{\sigma_k}(\boldsymbol{\theta}_k-\boldsymbol{\theta}_{k_i})
\end{equation}
\begin{equation}\label{eqcondF}
f(\boldsymbol{\theta}_k|\boldsymbol{\theta}_l)=\frac{\sum_{i=1}^{B}K_{\sigma_k}(\boldsymbol{\theta}_k-\boldsymbol{\theta}_{k_i})K_{\sigma_l}(\boldsymbol{\theta}_l-\boldsymbol{\theta}_{l_i})} {\sum_{i=1}^{B}K_{\sigma_l}(\boldsymbol{\theta}_l-\boldsymbol{\theta}_{l_i})}
\end{equation}
where $B$ is the buffer length and indicates the number of memory units which are allocated to save the history of the parameter, and $\boldsymbol{\theta}_{k_i}=\{\boldsymbol{\theta}_{k,i}, i\in\{n-1,n-2,...,n-B\} \}$.
Considering linear model\eqref{eqModel}, the likelihood function $f(D_{k,n}|\boldsymbol{\theta}_k) $ is defined as
\begin{equation} \label{eqLikelihoodData} 
\begin{split}
f(D_{k,n}|\boldsymbol{\theta}_k)&=f\left({\boldsymbol{d}}_{l} (n)|{\boldsymbol{u}}_{l,n} ,\boldsymbol{\theta}_{k,n} \right)\\
&\buildrel\Delta\over=K_{h_k}(L_\delta(e_{l,k}))\\
\end{split}
\end{equation}
where $\boldsymbol e_{l,k}=\boldsymbol{d}_l(n)-\boldsymbol{u}_{l,n}\boldsymbol{\theta}_k$
, and $L_\delta(a)$ is a pseudo-Huber loss function \cite{hartley2003} which is more robust against different noise, especially non-Gaussian one \cite{ashkezari2019}.  The pseudo-Huber loss function is defined as
\begin{equation} \label{eqsmHuber}
L_\delta(a)=\delta ^{2}\left({\sqrt {1+(a/\delta )^{2}}}-1\right).
\end{equation}
This function combines  $L_1$ squared loss 
and $L_2$ absolute loss by being convex when close to the minimum and less steep for extreme values. The steep can be controlled by $\delta$.
Taking the derivative of \eqref{eqLogLocalObjFun} with respect to $\boldsymbol{\theta}_k$ yields to
\begin{equation} \label{eqGradiantJhat}
\begin{split}
\nabla \hat{J_k}&=\sum_{l\in N_k}\frac{f'(D_l|\boldsymbol{\theta}_k)}{f(D_l|\boldsymbol{\theta}_k)}+\sum_{l\in N_k\setminus k}\left\{\frac{f'(\boldsymbol{\theta}_k|\boldsymbol{\theta}_l)}{f(\boldsymbol{\theta}_k|\boldsymbol{\theta}_l)}-\frac{f'(\boldsymbol{\theta}_k)}{f(\boldsymbol{\theta}_k)} \right\}\\
&=\frac{-1}{h_k}\sum_{l\in N_k}\frac{e_{l,k}(n)\boldsymbol{u}_{l,n}^{T}}{\sqrt{1+({e_{l,k}(n)}/{\delta})^2}}+\frac{1}{\sigma_k}\sum_{l\in N_k \setminus k}\sum_{i=1}^{B}\boldsymbol{\theta}_{k_i}(\boldsymbol\mu_{kli}-\boldsymbol\mu_{ki})
\end{split}
\end{equation}
where using kernel density estimation and its conditional KDE , $\mu_{kli}$ and $\mu_{ki}$ are defined as
\begin{equation}\label{eqMu_kli}
\boldsymbol\mu_{kli}=\frac{K_{\sigma_k}(\boldsymbol{\theta}_k-\boldsymbol{\theta}_{k_i})K_{\sigma_l}(\boldsymbol{\theta}_l-\boldsymbol{\theta}_{l_i})}{\sum_{j=1}^{B}K_{\sigma_k}(\boldsymbol{\theta}_k-\boldsymbol{\theta}_{k_j})K_{\sigma_l}(\boldsymbol{\theta}_l-\boldsymbol{\theta}_{l_j}) }
\end{equation}

\begin{equation}\label{eqMu_ki}
\boldsymbol\mu_{ki}=\frac{K_{\sigma_k}(\boldsymbol{\theta}_k-\boldsymbol{\theta}_{k_i})}{\sum_{j=1}^{B}K_{\sigma_k}(\boldsymbol{\theta}_k-\boldsymbol{\theta}_{k_j}) }
\end{equation}
\begin{figure}
	\centering
	\includegraphics[scale=.3]{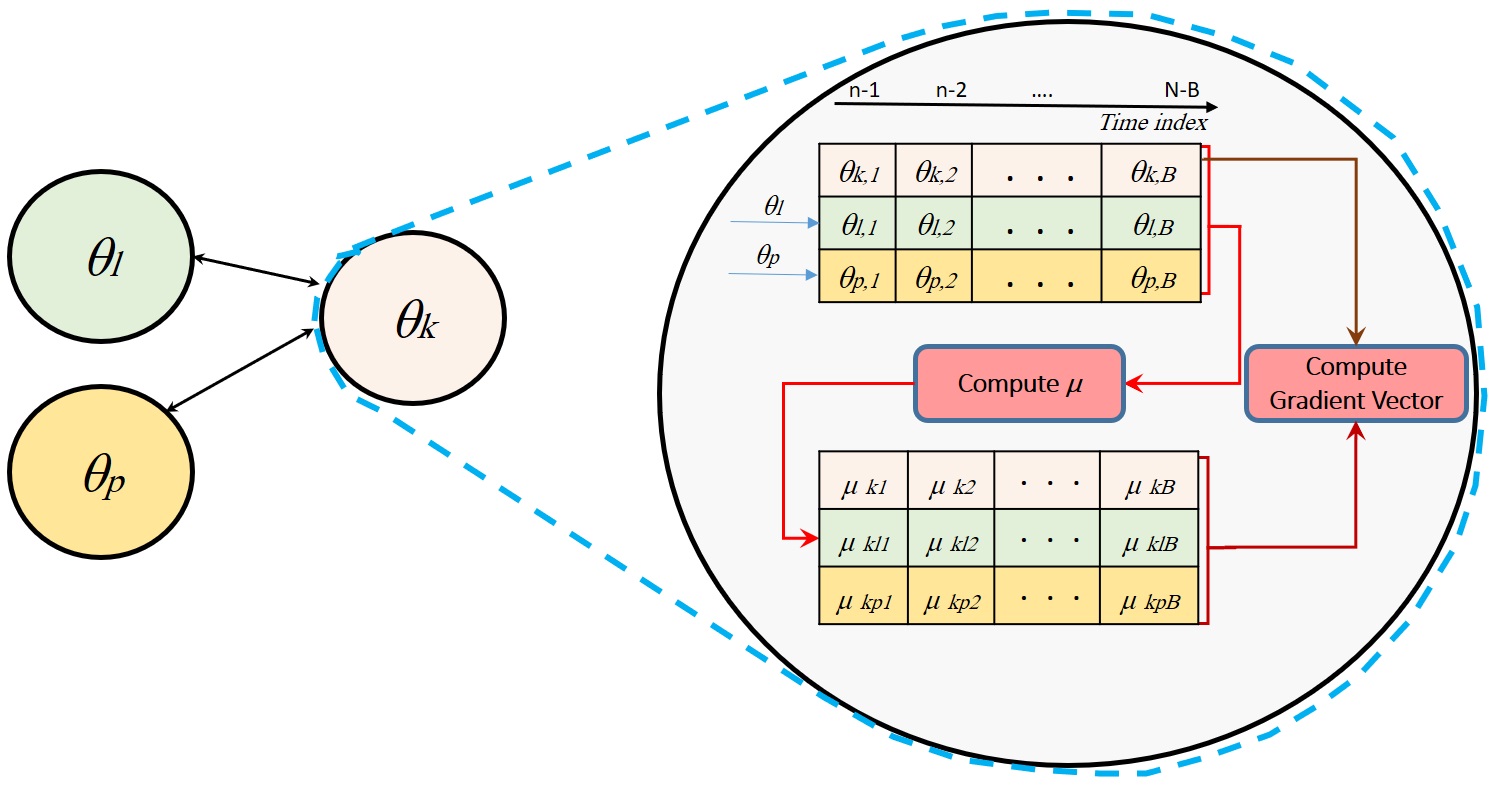}
	\caption{Sharing information of neighbors with $S_k$ . $S_k$ buffers the last $‌B$ received estimations of neighboring estimators to compute their joint weight considering $S_k$.}
	\label{fig_computeMu2}
\end{figure}

‌Defining the diagonal matrix $\beta_{k,i}$ we have
\begin{equation}\label{eqApproximateTheta}
\boldsymbol{\theta}_{k,i} =\beta_{k,i}\times \boldsymbol{\theta}_{k,n} 
\end{equation}

by substituting \eqref{eqApproximateTheta} in \eqref{eqGradiantJhat}, we have
\begin{equation} \label{eqGradiantJhat2}
\nabla \hat{J_k}=\frac{-1}{h_k}\sum_{l\in N_k}\frac{e_{l,k}(n)\boldsymbol{u}_{l,n}^{T}}{\sqrt{1+({e_{l,k}(n)}/{\delta})^2}}+\frac{1}{\sigma_k}\sum_{l\in N_k \setminus k}\sum_{i=1}^{B}\beta_{k,i}\boldsymbol{\theta}_{k,n} (\boldsymbol\mu_{kli}-\boldsymbol\mu_{ki})
\end{equation}
Finally, according to the order of adaption and combination steps, there are Adapt-Then-Combine (ATC) and Combine-Then-Adapt (CTA) strategies as shown in \eqref{eq16} and \eqref{eq17} respectively.
\begin{equation} \label{eq16} 
ATC \begin{cases}
{\boldsymbol{\phi}_{k,n} =\boldsymbol{\theta}_{k,n-1} -\alpha _{k}  \nabla \hat{J}_{k} \left(\boldsymbol{\theta}_{k,n-1} \right)} \\
{\boldsymbol{\theta}_{k,n} =\sum _{l\in N_{k} }a_{l,k} \boldsymbol{\phi}_{l,n} }
\end{cases}
\end{equation}

\begin{equation} \label{eq17} 
CTA \begin{cases}
{\boldsymbol{\phi}_{k,n-1} =\sum _{l\in N_{k} }a_{l,k} \boldsymbol{\theta}_{l,n-1}} \\ {\boldsymbol{\theta}_{k,n} =\boldsymbol{\phi}_{k,n-1} -\alpha _{k}  \nabla \hat{J}_{k} \left(\boldsymbol{\phi}_{k,n-1} \right)}
\end{cases}
\end{equation} 
where $\boldsymbol{\phi}$ is an intermediate estimation of $\theta_{o}$, $\alpha _{k} $ is the learning rate of the algorithm in $S_{k} $ and a trade-off between the speed of convergence and the steady-state error. The non-negative coefficients $a_{l,k}$ are elements of the left stochastic matrix $A=[a_{l,k} ]$, where satisfy 
\begin{equation}
a_{l,k}=0 \text{ if } l\notin N_k \text{,  }\sum_{l\in N_k} a_{l,k}=1, \ \ 
\end{equation}

\section{Performance analysis}
This section investigates the performance analysis of the proposed algorithm. Allow the local weight-error vectors be defined as
\begin{equation} \label{eqTildeTheta_k} 
\tilde{\boldsymbol{\theta}}_{k,n} ={\theta}_{o} -\boldsymbol{\theta}_{k,n}  
\end{equation} 
and form global weight-error vectors by stacking the local error vectors, i.e.
\begin{equation} \label{eqTildeTheta_n} 
\tilde{\boldsymbol{\theta}}_{n} \buildrel\Delta\over= col\left\{\tilde{\boldsymbol{\theta}}_{1,n} ,\tilde{\boldsymbol{\theta}}_{2,n} ,...,\tilde{\boldsymbol{\theta}}_{N,n} \right\}. 
\end{equation} 
Before analyzing the performance of NPDLMS, an approximation of the Gradient vector \eqref{eqGradiantJhat2} based on the Maclaurin series is computed. Here, just two first terms of the series are considered. Therefore, one has
\begin{equation}\
\frac{1}{\sqrt{1+({\boldsymbol e_{l,k}(n)}/{\delta})^2}}\simeq1-\frac{1}{2\delta^2}\boldsymbol e_{l,k}^2(n)
\end{equation} 
and, \eqref{eqGradiantJhat2} is approximated by
\begin{equation}\label{eqMclurnGradAprx}
\nabla {J}_{k}^{local} (\boldsymbol{\theta})\simeq -\frac{1}{h_k}\boldsymbol{u}_{l,n}^{T}e_{l,k}(n)+\frac{1}{2\delta^2h_k}\boldsymbol{u}_{l,n}^{T}\boldsymbol e_{l,k}^3(n)
+\frac{1}{\sigma_k}\sum_{l\in N_k \setminus k}\sum_{i=1}^{B}\beta_{k,i}\boldsymbol{\theta}_{k,n} (\boldsymbol\mu_{kli}-\boldsymbol\mu_{ki})
\end{equation}
When $n$ goes to infinity, the value of $\boldsymbol e_{l,k}(n)$ tends to zero and $\boldsymbol e_{l,k}^3(n)\le\boldsymbol e_{l,k}(n)$. Consequently,  \eqref{eqMclurnGradAprx} will be reduced to
\begin{equation}\label{eqMclurnGradAprxReduced}
\nabla {J}_{k}^{local} (\boldsymbol{\theta})\simeq \frac{1}{h_k}\boldsymbol{u}_{l,n}^{T}e_{l,k}(n)\left( \frac{1}{2\delta^2}-1\right)
+\frac{1}{\sigma_k}\sum_{l\in N_k \setminus k}\sum_{i=1}^{B}\beta_{k,i}\boldsymbol{\theta}_{k,n} (\boldsymbol\mu_{kli}-\boldsymbol\mu_{ki})
\end{equation}
Substituting $\boldsymbol{\phi}_{k,n-1} $ in $\boldsymbol{\theta}_{k,n} $ and also \eqref{eqMclurnGradAprxReduced} in \eqref{eq17} for CTA strategy leads to 
\begin{equation}\label{eqTheta_kReduced}
\begin{split}
\boldsymbol{\theta}_{k,n} =&\sum _{l\in N_{k} }a_{l,k} \boldsymbol{\theta}_{l,n-1} -\frac{\alpha _{k}}{h_k}  \left( \frac{1}{2\delta^2}-1\right)\boldsymbol{u}_{l,n}^{T}e_{l,k}(n)\\
&-\frac{\alpha _{k}}{\sigma_k}\sum_{l'\in N_k }a_{l',k} \boldsymbol{\theta}_{l',n-1}\sum_{l\in N_k \setminus k}\sum_{i=1}^{B}\beta_{k,i}  (\boldsymbol\mu_{kli}-\boldsymbol\mu_{ki})
\end{split}
 \end{equation}
 

Also defined are the following vectors and block matrices:
\begin{equation} \label{eq26} 
\boldsymbol{\it {\mathcal R}}_{n} \buildrel\Delta\over= diag\left\{\frac{2\delta^2-1}{2\delta^2h_k}\sum _{l\in N_{k} }\boldsymbol{u}_{l,n}^T \boldsymbol{u}_{l,n} ,k=1,..,N\right\} 
\end{equation} 
\begin{equation} \label{eq27} 
\boldsymbol{\it {\mathcal G}}_{n} \buildrel\Delta\over= col\left\{ \frac{2\delta^2-1}{2\delta^2h_k} \sum_{l\in N_{k} }{\boldsymbol{u}}_{1,n}^{T} {\boldsymbol{v}}_{1} (n), k= 1,...,N \right\} 
\end{equation} 

\begin{equation} \label{eqM} 
{\it {\mathcal M}}\buildrel\Delta\over= diag\left\{\alpha _{1} I_{d} ,\alpha _{2} I_{d} ,...,\alpha _{N} I_{d} \right\} 
\end{equation} 
\begin{equation} \label{eqP} 
\boldsymbol{\it {\mathcal P}}_{n} =diag\left\{\frac{1}{\sigma_k}\sum _{l\in N_{k}\setminus k }\sum_{i=1}^B\beta_{k,i}(\boldsymbol\mu_{kli}-\boldsymbol\mu_{ki}) ,k=1,...,N \right\}
\end{equation} 
Therefore, the global parameter vector is
\begin{equation}\label{eqTheta_n}
\boldsymbol{\theta}_n=\mathcal{A}\boldsymbol{\theta}_{n-1}-\mathcal M \boldsymbol{\mathcal R}_n\theta_o-\mathcal M \boldsymbol{\mathcal G}_n+\mathcal M \boldsymbol{\mathcal R}_{n-1}\mathcal{A}\boldsymbol{\theta}_{n-1}-\mathcal M \boldsymbol{\mathcal P}_n\mathcal{A}\boldsymbol{\theta}_{n-1}
\end{equation}
Now, subtracting both sides of (\ref{eqTheta_n}) from $\theta_{o} $ and using \eqref{eq26}-\eqref{eqP} and assuming $\mathcal{A}\theta_o=\theta_o$, the global error vector can be computed as 
\begin{equation} \label{eqErrVector} 
\tilde{\boldsymbol{\theta}}_{n} ={\boldsymbol {\mathcal F}}_{n} \tilde{\boldsymbol{\theta}}_{n-1} +{\it {\mathcal M}}\boldsymbol{\it {\mathcal G}}_n+\mathcal{M}\boldsymbol{\mathcal P}_n\theta_o
\end{equation} 
where ${\boldsymbol {\mathcal F}}_{n} =\left(I+{\it {\mathcal M}}\boldsymbol {\it {\mathcal R}}_n -{\it {\mathcal M}}\boldsymbol{\it {\mathcal P}}_{n}\right)\mathcal{A}$.

\subsection{ Mean convergence and stability}

To investigate the mean convergence and stability of the proposed algorithm, a recursion for the evolution of the network mean error vector is obtained taking the expectation of both sides of \eqref{eqErrVector} under Assumptions \ref{ass1}, \ref{ass2}, and \ref{ass3}
\begin{equation} \label{eqExpectationErr} 
\mathbb{E}\left(\tilde{\boldsymbol{\theta}}_{n} \right)={\it {\mathcal F}}{\it \mathbb{E}}\left(\tilde{\boldsymbol{\theta}}_{n-1} \right)+{\rm {\mathcal M}}{\rm {\mathcal G}}+\mathcal{M}\mathcal{P}\theta_o
\end{equation} 
where ${\it {\mathcal F}}=\mathbb{E}\left({\boldsymbol {\mathcal F}}_{n} \right)$, ${\rm {\mathcal G}}=\mathbb{E}\left(\boldsymbol{\it {\mathcal G}}_{n} \right)$ and $\mathcal P=\mathbb{E}(\boldsymbol{\it {\mathcal P}}_{n})$. Under Assumption \ref{ass3}, ${\rm {\mathcal G}}=0$.
Also,  
\begin{equation}
\begin{split}
\mathcal P=diag\left\{\frac{1}{\sigma_k}\sum _{l\in N_{k}\setminus k }\sum_{i=1}^B\beta_{k,i}\mathbb{E}(\boldsymbol\mu_{kli}-\boldsymbol\mu_{ki}) ,k=1,...,N \right\}
\end{split}
\end{equation}
Considering appendix equations \eqref{eqMu_kliNonEq},\eqref{eqMu_kiNonEq}, it can be written that 
\begin{equation}
\mathbb{E}(\boldsymbol\mu_{kli}-\boldsymbol\mu_{ki})\approx \left(\frac{1}{r_l}-\frac{1}{B}\right)=\frac{B-r_l}{Br_l}
\end{equation}
where $r_l$ is the number of buffered ${\boldsymbol{\theta}}_{l_j}$ which are similar to $\boldsymbol{\theta}_{l}$. Therefore,
\begin{equation}
\mathcal P\approx diag\left\{\frac{1}{\sigma_k}\sum _{l\in N_{k}\setminus k}\sum_{i=1}^B\beta_{k,i}\left(\frac{B-r_l}{Br_l}\right) ,k=1,...,N \right\}.
\end{equation}
\begin{equation}
\begin{split}
{\it {\mathcal F}}&=\mathbb{E} \left[ \left(I+{\it {\mathcal M}}\boldsymbol {\it {\mathcal R}}_n -\mathcal{M}\boldsymbol{\it {\mathcal P}}_{n}\right)\mathcal{A} \right]\\
&=\left(  I+{\it {\mathcal M}}{\rm {\mathcal R}}-{\it {\mathcal M}}{\mathcal P}\right)\mathcal{A}
\end{split}
\end{equation}
where $\mathcal R=\mathbb{E}\left(\boldsymbol{\mathcal R}_n\right)$.
Since $\mathcal P$ in \eqref{eqExpectationErr} is bounded, the algorithm will converge in the mean if $\mathcal{F}$ is a stable matrix, i.e.  $\mathop{\lim }\limits_{n\to \infty } \mathbb{E}\left(\tilde{\boldsymbol{\theta}}_{n+1} \right)=0$ if  $\rho \left({\it {\mathcal F}}\right)<1$, where $\rho \left({\it {\mathcal F}}\right)$ is the spectral radius of ${\it {\mathcal F}}$. This means that all eigenvalues of ${\it {\mathcal F}}$ are inside the unit circle. Since $\rho \left(A\right)=1$,
\[\rho \left({\it {\mathcal F}}\right)\le\rho \left(I+{\it {\mathcal M}}{\rm {\mathcal R}}-{\it {\mathcal M}}{\mathcal P}\right).\]
Choosing the learning rate for all nodes $S_k, k=1,...,N$ according to
\begin{equation} \label{eqConvergAlpha} 
0<\alpha _{k} <\frac{2}{\lambda_{\max }\left(\frac{1-2\delta^2}{2\delta^2h_k}\sum _{l\in N_{k} }{R}_{l}+\frac{1}{\sigma_k}\sum _{l\in N_{k}\setminus k}\sum_{i=1}^B\beta_{k,i}\left(\frac{B-r_l}{Br_l}\right)\right)}
\end{equation} 
where, ${R}_{l}=\mathbb{ E}(\boldsymbol{u}_{l,n}^T\boldsymbol{u}_{l,n})$ 
for each node will guarantee $\rho \left({\it {\mathcal F}}\right)<1$.

\begin{remark}
It is worth noting that for diffusion LMS \cite{R2}, the convergence condition is $0<\alpha _{k} <\frac{2}{\lambda _{\max } \left(\sum _{l\in N_{k} }{R}_{l} \right)} $. For conventional LMS, it is $0<\alpha <\frac{1}{\lambda _{\max } \left(R_{u} \right)} $ \cite{R23}.
\end{remark}

\subsection{ Mean square convergence and stability}

To study the mean-square performance of the proposed algorithms, $\mathbb{E}\left(\left\| \tilde{\boldsymbol{\theta}}_{n} \right\| _{\Sigma }^{2} \right)$ must be evaluated on \eqref{eqErrVector}, where $\Sigma $ is a positive semi-definite Hermitian matrix that is free to choose. Therefore,
\begin{equation} \label{eq38} 
\begin{split}
\mathbb{E}\left(\left\| \tilde{\boldsymbol{\theta}}_{n} \right\| _{\Sigma }^{2} \right)&={\it \mathbb{E}}\left(\left\| \tilde{\boldsymbol{\theta}}_{n-1} \right\| _{{\boldsymbol \Sigma }'}^{2} \right)
+{\boldsymbol{ \mathbb{E}}}\left(\boldsymbol{\it {\mathcal G}}_{n}^{T} {\it {\mathcal M}}\Sigma {\it {\mathcal M}}\boldsymbol{\it {\mathcal G}}_{n} \right)
\end{split}+\mathbb{E}\left(\theta_o^T\boldsymbol{\mathcal{P}}_n^T\mathcal{M}\Sigma\mathcal{M}\boldsymbol{\mathcal{P}}_n\theta_o\right)
\end{equation} 

In this equation,
\begin{equation} \label{eq39} 
{\boldsymbol \Sigma }'={\boldsymbol {\mathcal F}}_{n}^{T} \Sigma {\boldsymbol {\mathcal F}}_{n}  .                     
\end{equation} 

Also, the expectation of ${\boldsymbol \Sigma }'$ is denoted as
\begin{equation} \label{eq40} 
{\rm \Sigma }'=\mathbb{E}\left({\boldsymbol \Sigma }'\right).                   
\end{equation} 

Assuming $\tilde{\boldsymbol{\theta}}_{n} $, $\boldsymbol{\it {\mathcal R}}_{n} $, and $\boldsymbol{\it {\mathcal P}}_{n} $ are independent, so that $\mathbb{E}\left(\left\| \tilde{\boldsymbol{\theta}}_{n-1} \right\| _{{\boldsymbol \Sigma }'}^{2} \right)={\bf \mathbb{E}}\left(\left\| \tilde{{\boldsymbol \theta }}_{n-1} \right\| _{{\rm \Sigma }'}^{2} \right)$, by employing \eqref{eq40}.
\\ \\
\textbf{Fact 1}: For any matrices ${\rm {\rm A} ,\; \Sigma }$, and $B$ of appropriate sizes, the following holds \cite{R24}:\\
a) $vec\left(A{\rm \Sigma {\rm B} }\right)=\left(B^{T} \otimes A\right)vec\left({\rm \Sigma }\right)$~~~~
b) $Tr\left(A^{T} B\right)=vec\left(A\right)^{T} vec\left(B\right)$ \\
Let $\sigma \buildrel\Delta\over= vec\left(\Sigma \right)$ and $\sigma '\buildrel\Delta\over= vec\left({\rm \Sigma }'\right)$. Also, the notation $\left\| \tilde{\boldsymbol{\theta}}\right\| _{\sigma }^{2} $ is employed to denote $\left\| \tilde{\boldsymbol{\theta}}\right\| _{{\rm \Sigma }}^{2} $. Using Fact 1(a), \eqref{eq39}, and \eqref{eq40}, ${\rm \Sigma }'$ can be vectorized as follows
\begin{equation} \label{eq41} 
vec(\Sigma ')=vec\left({\rm {\mathcal F}}^{T} \Sigma {\rm {\mathcal F}}\right)=\left({\rm {\mathcal F}}^{T} \otimes {\rm {\mathcal F}}^{T} \right)vec\left(\Sigma \right).            
\end{equation} 
Using 
\begin{equation} \label{eq42} 
F={\rm {\mathcal F}}^{T} \otimes {\rm {\mathcal F}}^{T}  ,                                       
\end{equation} 
then 
\begin{equation} \label{eq43} 
\sigma '=F\sigma .                           
\end{equation} 
Using Fact 1, one has following equations:
\begin{equation} \label{eq44} 
\begin{split}
\mathbb{E}\left(\boldsymbol{\it {\mathcal G}}_{n}^{T} {\it {\mathcal M}}\Sigma {\it {\mathcal M}}\boldsymbol{\it {\mathcal G}}_{n} \right)&=Tr\left({\it \mathbb{E}}\left(\boldsymbol{\it {\mathcal G}}_{n} \boldsymbol{\it {\mathcal G}}_{n}^{T}  {\it {\mathcal M}}\Sigma {\it {\mathcal M}}\right)\right)\\
&=vec(G)^{T} vec\left({\it {\mathcal M}}\Sigma {\it {\mathcal M}}\right)\\
&=\gamma \sigma
\end{split}
\end{equation} 
where $G^T=\mathbb{E}\left(\boldsymbol{\it {\mathcal G}}_{n}\boldsymbol{\it {\mathcal G}}_{n}^{T}  \right)$ and $\gamma =vec\left(G \right)^{T} \left({\it {\mathcal M}}\otimes {\it {\mathcal M}}\right)$.
\begin{equation}\label{eq441}
\begin{split}
\mathbb{E}\left(\theta_o^T\boldsymbol{\mathcal{P}}_n^T\mathcal{M}\Sigma\mathcal{M}\boldsymbol{\mathcal{P}}_n\theta_o\right)&=Tr(\mathbb{ E}\left(\boldsymbol{\mathcal{P}}_n\theta_o\theta_o^T\boldsymbol{\mathcal{P}}_n^T\mathcal{M}\Sigma\mathcal{M}\right))\\
&=vec(P)^Tvec(\mathcal{M}\Sigma\mathcal{M})\\
&=\delta\sigma
\end{split}
\end{equation}
where $P=\mathbb{E}\left(\boldsymbol{\mathcal{P}}_n\theta_o\theta_o^T\boldsymbol{\mathcal{P}}_n^T\right)$ and $\delta =vec\left(P \right)^{T} \left({\it {\mathcal M}}\otimes {\it {\mathcal M}}\right)$.
Thus, substituting \eqref{eq44} and \eqref{eq441} in \eqref{eq38} and considering \eqref{eq43} this can be rewritten as

\begin{equation} \label{eq47} 
\mathbb{E}\left(\left\| \tilde{\boldsymbol{\theta}}_{n} \right\| _{\sigma }^{2} \right)={\it \mathbb{E}}\left(\left\| \tilde{\boldsymbol{\theta}}_{n-1} \right\| _{F\sigma }^{2} \right)+\gamma \sigma  +\delta\sigma
\end{equation} 
To show the mean-square stability one can write
\begin{equation} \label{eq48} 
\mathop{\lim }\limits_{n\to \infty } \mathbb{E}\left(\left\| \tilde{\boldsymbol{\theta}}_{n} \right\| _{\sigma }^{2} \right)=\mathop{\lim }\limits_{n\to \infty } \left\{{\it \mathbb{E}}\left(\left\| \tilde{\boldsymbol{\theta}}_{n-1} \right\| _{F\sigma }^{2} \right)+\gamma \sigma+\delta\sigma \right\}.         
\end{equation} 
Therefore, the stability of the proposed algorithm depends on the stability of $F$. According to \eqref{eq42}, $F$ will be stable if ${\rm {\mathcal F}}$ is stable. According to mean-square stability analysis, the stability of \eqref{eq48} is guaranteed if \eqref{eqConvergAlpha} holds. Thus \eqref{eqConvergAlpha} is sufficient to guarantee the mean and mean-square stability.

\subsection{ Mean-square steady state analysis}
Now the mean-square steady-state performance for the network is considered. Letting $n\to \infty $ and using \eqref{eq47},

\begin{equation} \label{eq49} 
\mathop{\lim }\limits_{n\to \infty } \left\{\mathbb{E}\left\| \tilde{\boldsymbol\theta }_{n} \right\| _{(I-F)\sigma }^{2} \right\}=\xi \sigma .                  
\end{equation} 
where $\xi=\gamma+\delta$.
The MSD and EMSE quantities at each node $S_{k} $ are defined respectively as
\begin{equation} \label{eq50} 
\varsigma_{k} =\mathop{\lim }\limits_{n\to \infty } \left\{\mathbb{E}\left\| \tilde{\boldsymbol\theta }_{k,n} \right\| _{I}^{2} \right\},              
\end{equation} 
\begin{equation} \label{eq51} 
\varrho_{k} =\mathop{\lim }\limits_{n\to \infty } \left\{\mathbb{E}\left\| \tilde{\boldsymbol\theta }_{k,n} \right\| _{R_{u,k}}^{2} \right\}.             
\end{equation} 
Thus the MSD at node $S_{k} $ is obtained by weighting $\left\| \tilde{\boldsymbol\theta }_{k,n} \right\| _{}^{2} $ with a block matrix that has an identity matrix at block $\left(k,k\right)$ and zeros elsewhere, i.e. $diag\left(\boldsymbol e_{k} \right)\otimes I_{M} $, when $n\to \infty $. Therefore assuming $\left(I-F\right)$ is invertible and using \eqref{eq49} and \eqref{eq50} MSD for each node is calculated by
\begin{equation} \label{eq52} 
\varsigma_{k} =\xi(I-F)^{-1} vec\left(diag\left(\boldsymbol e_{k} \right)\otimes I_{d} \right).           
\end{equation} 
where $F$ is given by \eqref{eq42}. Similarly, the EMSE at node $S_{k} $ is obtained by weighting $\left\| \tilde{\boldsymbol\theta }_{k,n} \right\| _{}^{2} $ with a block matrix that has $R_{u,k} $ at block $\left(k,k\right)$ and zeros elsewhere, i.e. $diag\left(\boldsymbol e_{k} \right)\otimes R_{u,k} $, when $n\to \infty $.  Then one can obtain
\begin{equation} \label{eq53} 
\varrho_{k} =\xi (I-F)^{-1} vec\left(diag\left(\boldsymbol e_{k} \right)\otimes R_{u,k} \right).       
\end{equation} 
The whole network MSD and EMSE are defined as the average of $\varsigma_{k} $ and $\varrho_{k} $ over all the nodes, i.e.
\begin{equation} \label{eq54} 
\varsigma=\frac{1}{N} \sum _{k=1}^{N}\varsigma_{k}  ,{\rm \; \; }\varrho=\frac{1}{N} \sum _{k=1}^{N}\varrho_{k}  .                  
\end{equation} 

\subsection{ Mean-square transient analysis}

To compute the instantaneous MSD and EMSE for every node $S_{k} $ and analysis the mean square behavior in the transient state, the following recursion is made based on \eqref{eq47} by considering $\boldsymbol{\theta}_{k,-1} =0,\forall k\in\{1,2,...,N\} $
\begin{equation} \label{eq154} 
\left\| \tilde{\boldsymbol{\theta}}_{n} \right\| _{\sigma }^{2} = \left\| {\theta}_{o} \right\| _{F^{n+1}\sigma }^{2}+\xi \sum_{j=0}^{n}F^j\sigma
\end{equation} 
Now writing this recursion for $n-1$, and subtracting it from \eqref{eq154} leads to
\begin{equation} \label{eq55} 
\left\| \tilde{\boldsymbol\theta }_{n} \right\| _{\sigma }^{2} =\left\| \tilde{\boldsymbol\theta }_{n-1} \right\| _{\sigma }^{2} +\left\| {\theta}_{o} \right\| _{F^{n} \left(I-F\right)\sigma }^{2} +\xi F^{n} \sigma  .               
\end{equation} 
Replacing $\sigma $ with $\sigma _{msd_k} =diag\left(\boldsymbol e_{k} \right)\otimes I_{M} $, $\sigma _{emse_k} =diag\left(\boldsymbol e_{k} \right)\otimes R_{u,k} $ in \eqref{eq55}, the instantaneous MSD and EMSE can be computed recursively over time
\begin{equation}\label{eq156}
\begin{split}
\varsigma(n) &= \varsigma(n-1)-\left\| {\theta}_{o} \right\| _{F^{n} \left(I-F\right)\sigma_{msd_k} }^{2} +\xi F^{n} \sigma_{msd_k}\\
\varrho(n) &= \varrho(n-1)-\left\| {\theta}_{o} \right\| _{F^{n-1} \left(I-F\right)\sigma_{emse_k} }^{2} +\xi F^{n-1} \sigma_{emse_k}.
\end{split}
\end{equation}
\section{Reducing the sensitivity of the algorithm and computational overhead}
Now, the neighbors error of node $S_k$ is defined as
\begin{equation}\label{eqEpsilon}
\epsilon_k=\sum_{l\in N_{k}}\left(\boldsymbol{d}_l-{\boldsymbol{u}}_{l} \boldsymbol{\theta}_{k}\right)^2 
\end{equation}

\begin{figure}
	\centering
	\includegraphics[scale=.45]{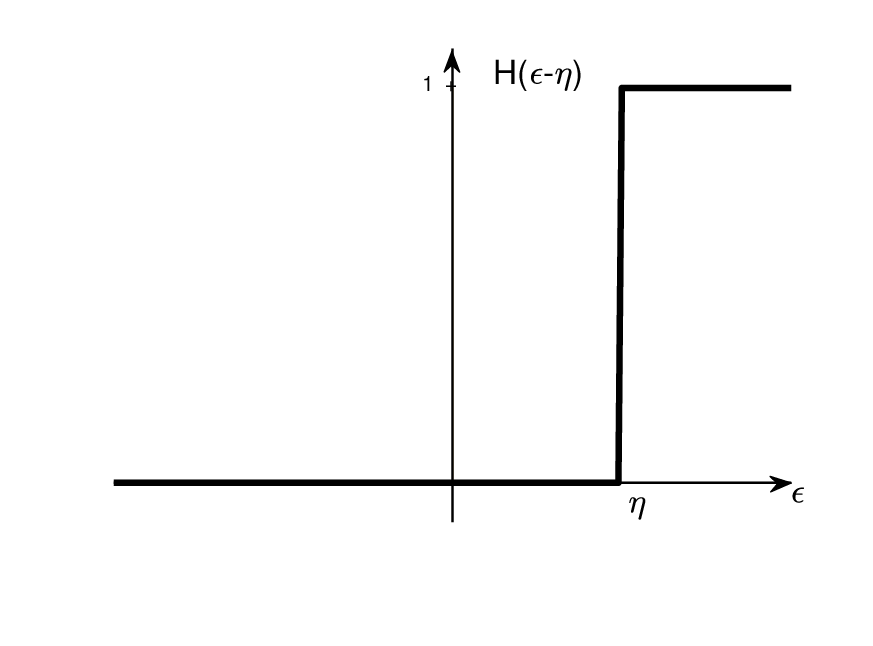}
	\caption{The error function H clips the error}
	\label{Hfun}
\end{figure}

To relax the update rule, the error function $H$ is defined as follow (see Fig. \ref{Hfun}), 
\begin{equation}\label{eqH}
H(\epsilon_k-\eta)=\frac{1}{1+exp(-2s(\epsilon_k-\eta))},
\end{equation}
where $s$ is a scalar and $\eta$ is a predefined threshold. Therefore, the update rule will be 
\begin{equation}\label{eqUpdateRule1}
\boldsymbol{\theta}_{k,n}=\boldsymbol{\theta}_{k,n-1}+\alpha_k H(\epsilon_k-\eta)\nabla \hat{J_k}
\end{equation} 
which is equal to 
\begin{equation}\label{eqUpdateRulIf}
\boldsymbol{\theta}_{k,n}=
\begin{cases}
\boldsymbol{\theta}_{k,n-1}-\alpha_k\nabla \hat{J_k} & \text{if $\epsilon_k>\eta$}.\\
\boldsymbol{\theta}_{k,n-1} & \text{otherwise}.
\end{cases}
\end{equation}
Therefore, considering suitable $\eta$ will help to reduce the computational time while preserving the desired accuracy. It has been discussed in the next section.

\section{Simulation Results}
\subsection{Comparing computational complexity}
Table 1 shows the computational complexity of the diffusion LMS \cite{R8}, the DMCC \cite{ma2016}, the DSE-LMS \cite{ni2016b}, the DLMS/F \cite{zheng2017}, and the DLLAD \cite{chen2018} algorithms for each node $k$ in the network. The number of multiplications, additions, absolute value and sign operator has been reported. As seen, the computational complexity of the proposed algorithm is greater than other algorithms because of the existence of a buffer to save the historical data.  But, it is important to note that the value of $‌B$ which indicates the buffer length is usually less than 5 in practice. The kernel size $\sigma$ in the DMCC  \cite{ma2016}, positive parameter $\lambda$ in  the DLMS/F \cite{zheng2017}, parameter $\alpha$ in DLLAD \cite{chen2018} and $h_k$, $\sigma_k$ and $\delta$ in \eqref{eqGradiantJhat2} are parameters which must be set correctly. 

\begin{table} \label{t1}
	\centering
	\caption{Computational complexity of different algorithms}
	\label{tableGradient}
	\begin{tabular}{p{2.25cm} p{4cm} p{4cm} p{0.75cm} p{0.75cm}}
		\hline
		Algorithm & $\times$&$+$&$|.|$&$\sign(.)$ \\ \hline \hline
		DLMS \cite{R8}&   $(|N_k|+2)d+1$& $(|N_k|+1)d$&0&0   \\ \hline
		DSE-LMS \cite{ni2016b}&  $(3d+1)|N_k|+1$&$(4d-1)|N_k|-d$ &0&$|N_k|$\\ \hline
		DLMS/F \cite{zheng2017}&   $ (4d+3)|N_K|+M$&$(2d+2)|N_k|$&0&0 \\ \hline
		DMCC \cite{ma2016}& $(|N_k|+2)d+1$&$(|N_k|+1)d$&0&0     \\ \hline
		DLLAD\cite{chen2018}&$(|N_k|+2)d+3$&$|N_k|d+2$&1&0\\ \hline
		NPDLMS\dag&$(dB+4‌B+3d+1)|N_k|+2d-‌B(d+4)$&$(2d+2+7dB‌+7‌‌B)|N_k|-7‌B(d+1)-d$&0&0 \\ \hline
		
	\end{tabular}
\begin{tabular}{p{0.5cm} p{12cm}}
\dag& The existence of a buffer with a length of $B$ has been caused to a more computational effort in the NPDLMS algorithm.
\end{tabular}
\end{table}

\subsection{Evaluating different algorithms in the system identification}
\begin{figure}[h]
	\centering
	\subfloat[]{
		\label{fig:network}
		\includegraphics[trim={0.001cm 0 0.001cm 0},clip,scale=.9]{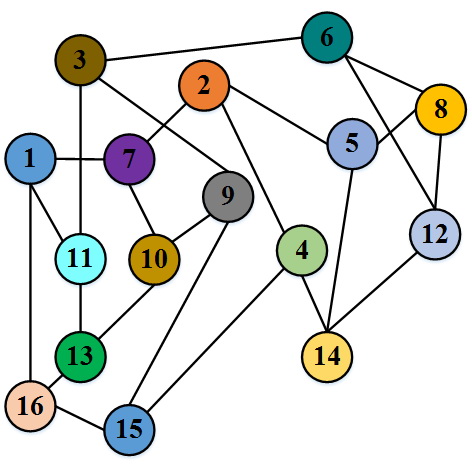}}
	\subfloat[]{
		\label{fig:networkprofile}
		\includegraphics[trim={.1cm 0 1.cm 0},clip,scale=.45]{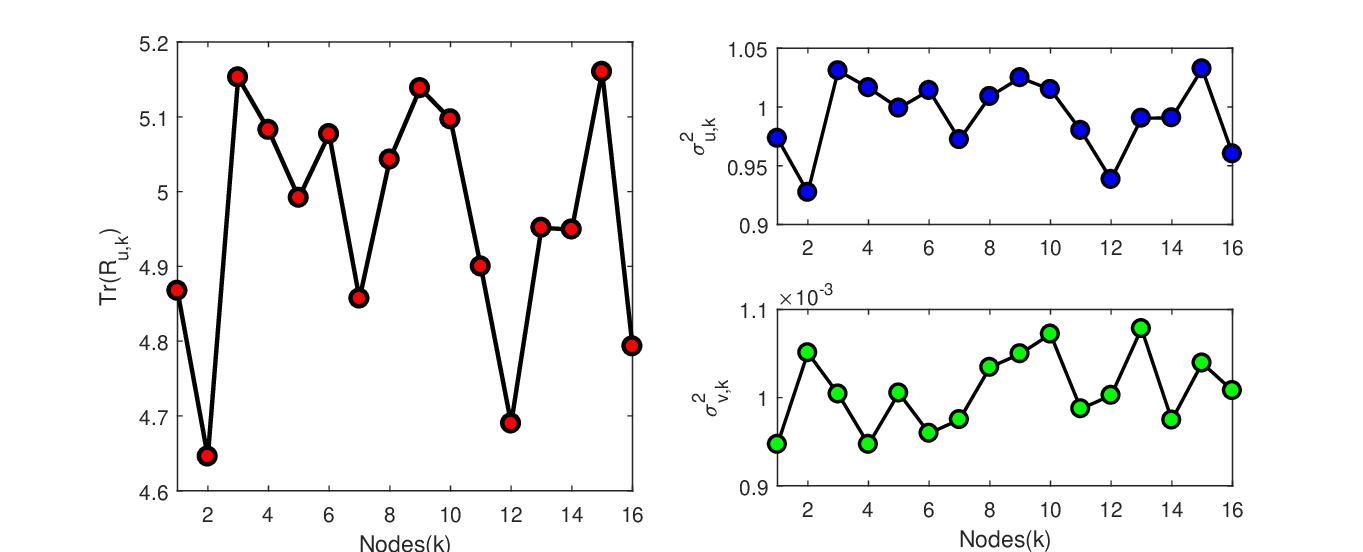}}
	\caption[stationaryNetwork]{Network setting:
		\subref{fig:network} Network topology,
		\subref{fig:networkprofile} Network variance profile for three different noise levels of the model.}
	\label{fig_stationaryNetwork}
\end{figure}

This section presents simulation results to analyze the efficiency of the proposed algorithm by comparing the learning curve of the MSD of the proposed algorithm along with the literature in the stationary and non-stationary environments. In the stationary environments, the unknown parameter vector is fixed over time while, in non-stationary environments, the unknown parameter vector varies over time. The CTA version of the currently proposed algorithm is compared with the CTA strategy of the diffusion LMS \cite{R8}, the DMCC \cite{ma2016}, the DSE-LMS \cite{ni2016b}, the DLMS/F \cite{zheng2017} and the DLLAD \cite{chen2018} algorithms. 
Consider a network with a set of sixteen nodes $V=\left\{S_{1} ,S_{2} ,...,S_{16} \right\}$ and set of edges as depicted in the topology of the network in Fig. \ref{fig:network}. The scalar measurements in the nodes are described according to model \eqref{eqModel}, ${\boldsymbol{d}}_{k} (n)={\boldsymbol{u}}_{k,n} {\theta}_{o} +{\boldsymbol{v}}_{k} (n), k=1,2,...,16$, where the regressors are size $d=5$, zero-mean Gaussian, independent in time and space, and have the covariance matrices $R_{u,k}$. The unknown parameter vector is set to ${\theta}_{o} =1_{5} /\sqrt{5} $. The variance of the model noise is denoted by $\sigma ^{2} _{v,k} $ and has the $SNR\triangleq10\log_{10}(\sigma^2_d/\sigma^2_{v,k})$. The network energy profile is shown in Fig. \ref{fig:networkprofile} for SNR = 30 dB. 
\subsubsection{Comparison of learning behavior in the stationary environment}
\begin{figure}[h]
	\centering
	\subfloat[]{
		\label{fig:stationarygaussNoise}
		\includegraphics[width=2.6in]{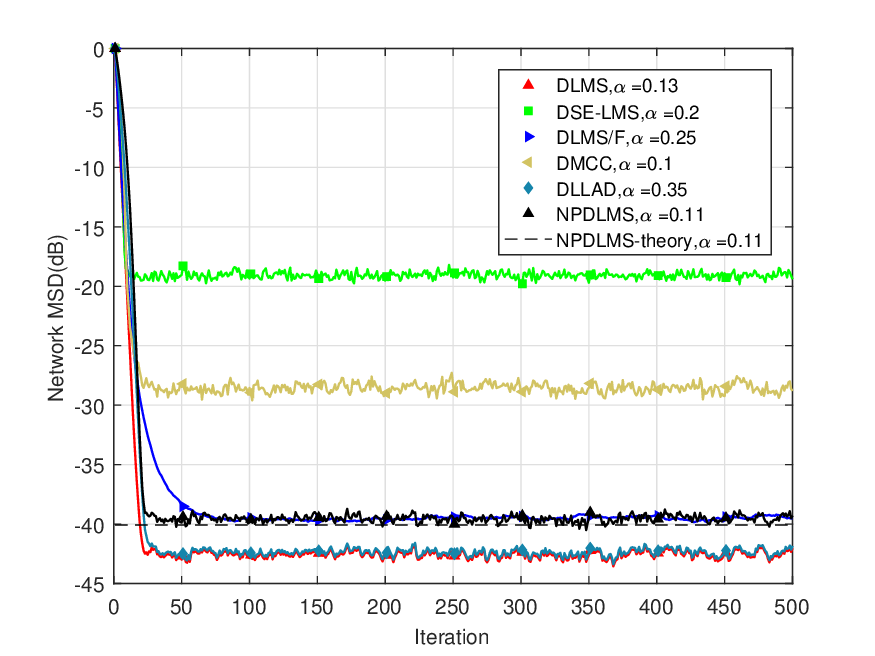}}
	\subfloat[]{
		\label{fig:stationarygaussMuNoiseNegSNR}
		\includegraphics[width=2.5in]{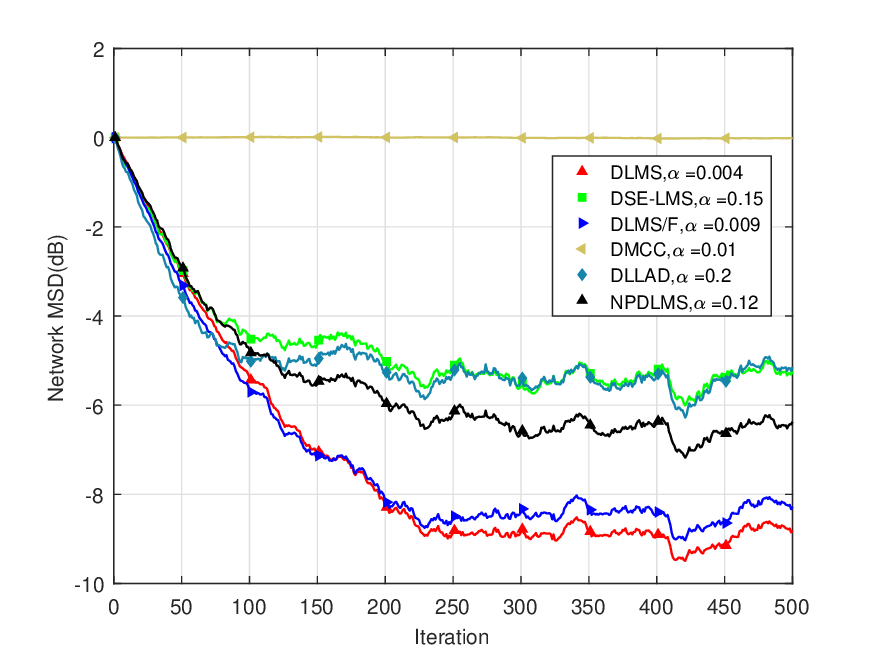}}
	\caption[stationaryGaussiannoise]{Comparison of algorithms behavior for Gaussian noise in a stationary environment:
		\subref{fig:stationarygaussNoise} Gaussian noise with SNR =$30$ dB;
		\subref{fig:stationarygaussMuNoiseNegSNR} Gaussian noise with SNR =$-20$ dB.}
	\label{fig:stationaryNetwork}
\end{figure}

The aim of this example is to identify the mathematical model of the unknown system from measured data  in a stationary environment. In this case,  statistical properties of the system are invariant.  
The system performance has been investigated by performing and averaging 200 independent experiments. Each curve was generated by running the learning process for 500 iterations. To obtain similar initial convergence, algorithms' step size has been set to $0.2$ for the DSE-LMS, $0.1$ for the DMCC, $0.25$ for the DLMS/F,  $0.13$ for the DLMS, $0.35$ for the DLLAD and $0.11$ for the NPDLMS when the SNR is $30$ dB. In SNR =$-20$ dB, the step-size is set to $0.15$ for the DSE-LMS, $0.009$ for the DLMS/F, $0.01$ for the DMCC, and $0.004$ for the DLMS, $0.2$ for the DLLAD and $0.12$ for the NPDLMS. 
Fig. \ref{fig:stationarygaussNoise} shows the MSD curves when the environment contaminated by Gaussain noise with SNR=$30$ dB, and Fig. \ref{fig:stationarygaussMuNoiseNegSNR} shows the result when SNR =$-20$ dB and $\delta=0.25$. The DLMS, DLMS/F, DLLAD and NPDLMS outperforms the DMCC and the DSE-LMS in both positive and negative SNRs. IN SNR=$30$ dB, DLMS/F and NPDLMS have similar steady-state but NPDLMS has a more rapid convergence rate. 

The non-Gaussian noise is modeled by a $\alpha$-stable distribution function $V_{\alpha-stable}(\alpha_v,\beta_v,\gamma_v,\delta_v)$. This model is widely utilized as an impulsive noise mode in the literature \cite{shao1993,ma2016}. The  characteristic function of $\alpha$-stable process is defined as
\begin{equation}\label{eqAlphaNoise}
\phi(t)= \exp\left( -\gamma_v |t|^{\alpha_v} [ 1 + i \beta_v sign(t) S(t,\alpha_v) ] + i \delta_v t \right)
\end{equation}
where
$S(t,\alpha_v)=
\begin{cases}
\tan(\frac{\pi \alpha_v}{2} ), &  \alpha_v \neq 1 \\
\frac{2}{\pi} \log|t|,  &  \alpha_v=1
\end{cases}$
, the parameter $\alpha_v \in (0,2]$ is called the characteristic exponent and describes the tail of the distribution,  $\beta_v \in [-1,1]$ is the skewness and specifies if the distribution is right ($\beta_v > 0$) or left ($\beta_v < 0$) skewed. The last two parameters are the scale, $\gamma_v > 0$, and the location $\delta_v \in R$ \footnote{Matlab implementation is available on http://math.bu.edu/people/mveillet/html/alphastablepub.html}.
\begin{figure} [h]
	\centering
	\includegraphics[width=2.6in]{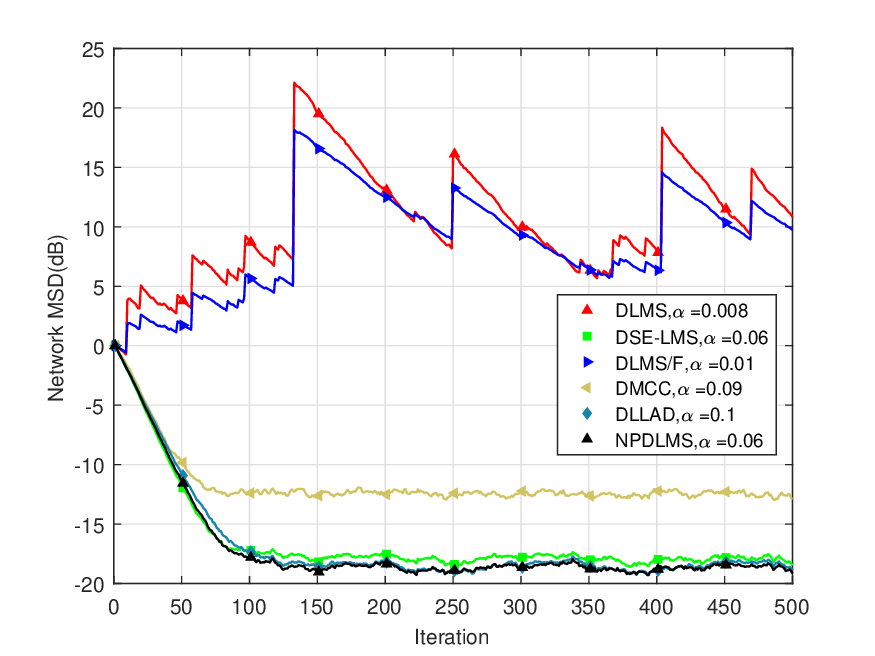}
	\caption[stationarynonGaussiannoise]{Comparison of algorithms behavior for non-Gaussian noise in a stationary environment. The non-Gaussian noise is $V_{\alpha-stable}(1.2,0,1,0)$}
	\label{fig:stationaryNongaussNoise}
\end{figure}
Fig. \ref{fig:stationaryNongaussNoise} shows the MSD curves of competitor algorithms. The non-Gaussian noise is $V_{\alpha-stable}(1.2,0,1,0)$. The step-size is set to $0.008$ for the DLMS, $0.01$ for the DLMS/F, $0.09$ for the DMCC, $0.06$ for the DSE-LMS, $0.1$ for the DLAAD and $0.06$ for the NPDLMS to obtain a same initial convergence rate.  As seen, the DLMS and DLMS/F has been diverged. The proposed NPDLMS and DSE-LMS algorithms outperform others. 
\subsubsection{Comparison of learning behavior in non-stationary environment}
\begin{figure}[!h]
	\centering
	\subfloat[]{
		\includegraphics[width=2.6in]{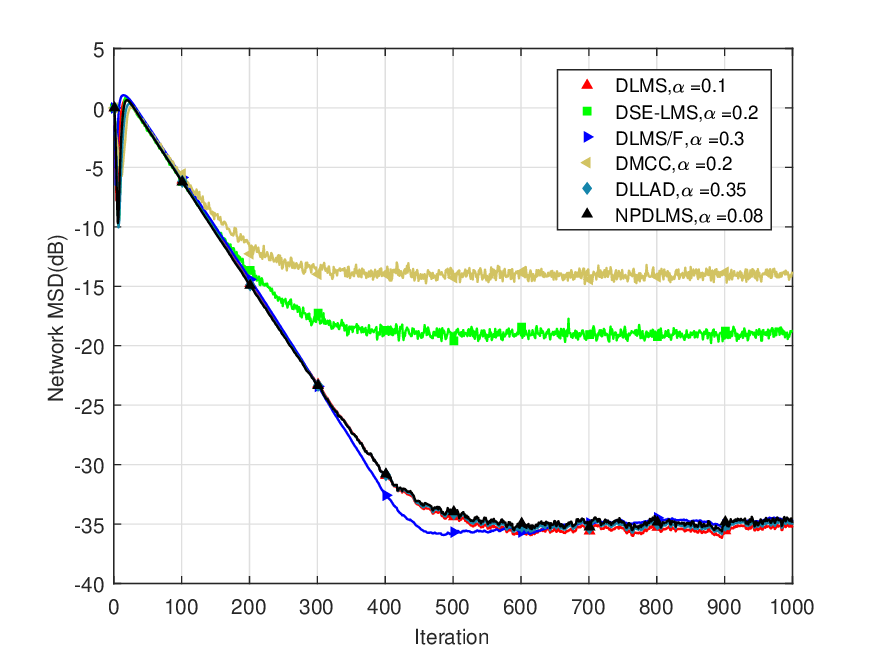}
		\label{fig:nonstationarygaussNoise}}
	\subfloat[]{
		\includegraphics[width=2.6in]{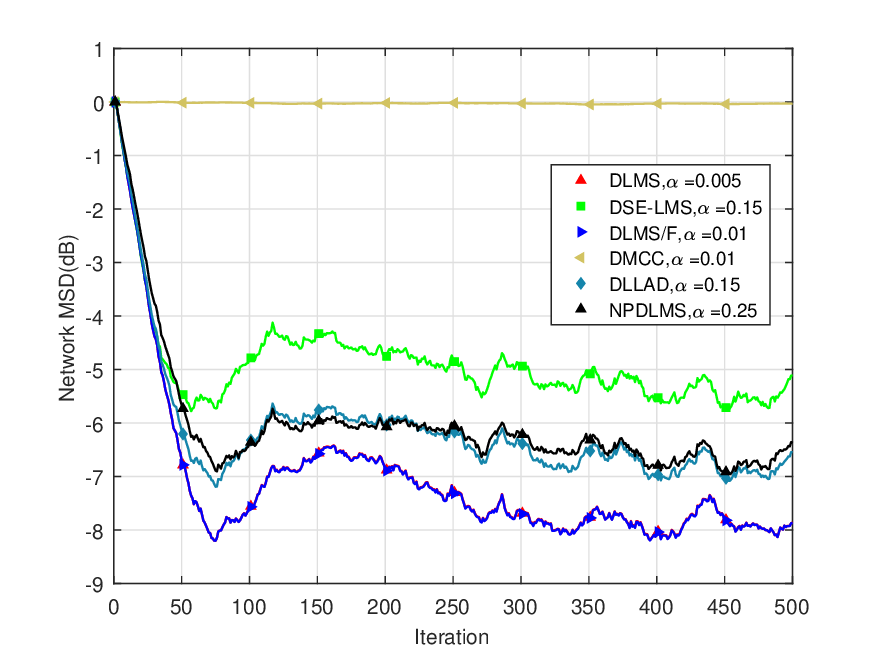}
		\label{fig:nonstationarygaussNoiseNegSNR}}
	\caption[nonstationarynonGaussiannoise]{Comparison of algorithms behavior for Gaussian noise in non-stationary environment:
		\subref{fig:nonstationarygaussNoise} Gaussian noise with SNR = $30$dB;
		\subref{fig:nonstationarygaussNoiseNegSNR} Gaussian noise with SNR = $-20$dB }
	\label{fig:nonstationaryNetwork}
\end{figure}

This example investigates the performance of the proposed algorithm in comparison with the literature in a non-stationary environment where statistical properties change during the time. 
In this case, the parameter vector of interest ${\theta}_{o} $ changes according to the tracking model \eqref{eq56} which is taken from \cite{R25},

\begin{equation} \label{eq56} 
M_{1} =\left\{\begin{array}{l} {\boldsymbol{\theta}_{o,n} ={\theta}_{o} +\omega {\rm }_{n} } \\ {\omega {\rm }_{n} =0.99\omega {\rm }_{n-1} +q_{n} } \end{array}\right.  
\end{equation} 
where $q_{n} $ is a sequence of i.i.d perturbations with a zero mean and covariance matrix $Q$, and is independent of the input regressors and output noise at every iteration. $Q=\sigma _{q}^{2} I$ is selected and the value of $\sigma _{q}^{2} $ is $\sigma _{q}^{2} =10^{-4} $. The model of measurement is similar to \eqref{eqModel}, where ${\boldsymbol{v}}_{k} (n)$ is an additive zero mean white noise. The network topology is the same as that shown in Fig. \ref{fig:network}. 
By averaging over 200 realizations over 1000 iterations, Fig. \ref{fig:nonstationarygaussNoise} shows the convergence and steady-state network MSD of the competitor algorithms for the Gaussian noise with SNR $30$ dB, and Fig. \ref{fig:nonstationarygaussNoiseNegSNR} shows the result for the Gaussian noise with SNR $-20$ dB . In the first case the step-size is set to 0.1 for the DLMS, 0.2 for the DES-LMS and the DMCC, 0.3 for the DLMS/F, 0.35 for the DLLAD and 0.15 for the NPDLMS. The DLMS, DLMS/F, DLAAD and NPDLMS outperforms other algorithms.
In the second case, the step-size is set to 0.005 for the DLMS , 0.15 for the DSE-LMS and DLAAD, 0.25 for the NPDLMSand 0.01 for the DLMS/F and DMCC. For Gaussian noise with SNR =  $-20$ dB, the proposed NPDLMS algorithm is similar to the DLAAD, better than the DSE-LMS and outperforms the DMCC while the DLMS and  DLMS/F algorithms outperforms all algorithm.   

\begin{figure} [h]
	\centering
	\includegraphics[width=2.6in]{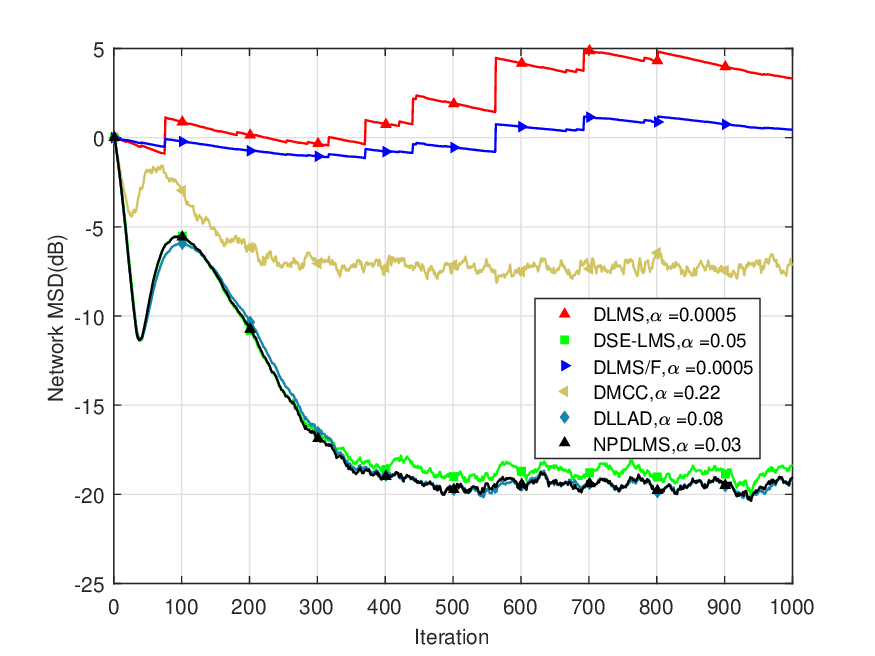}
	\caption{Comparison of algorithms behavior for the non-Gaussian noise $V_{\alpha-stable}(1.2,0,1,0)$ in the non-stationary environment}
	\label{fig:nonstationary_nongauss}
\end{figure}

In the presence of non-Gaussian noise, the noise model follows \eqref{eqAlphaNoise} and is modeled by $\alpha$-stable distribution $V_{\alpha-stable}(1.2,0,1,0)$. In this case
the step size is 0.0005 for the DLMS and DLMS/F, 0.05 for the DSE-LMS, 0.22 for the DMCC, 0.08 for the DLLAD, and 0.03 for the NPDLMS. As seen in Fig. \ref{fig:nonstationary_nongauss}, the proposed NPDLMS has a similar behavior with DSE-LMS and DLAAD while these algorithms outperform other algorithms.
\subsection{Investigating the steady state performance for non-Gaussian noise }
\begin{figure}[!h]
	\centering
	\subfloat[]{
		\includegraphics[width=2.6in]{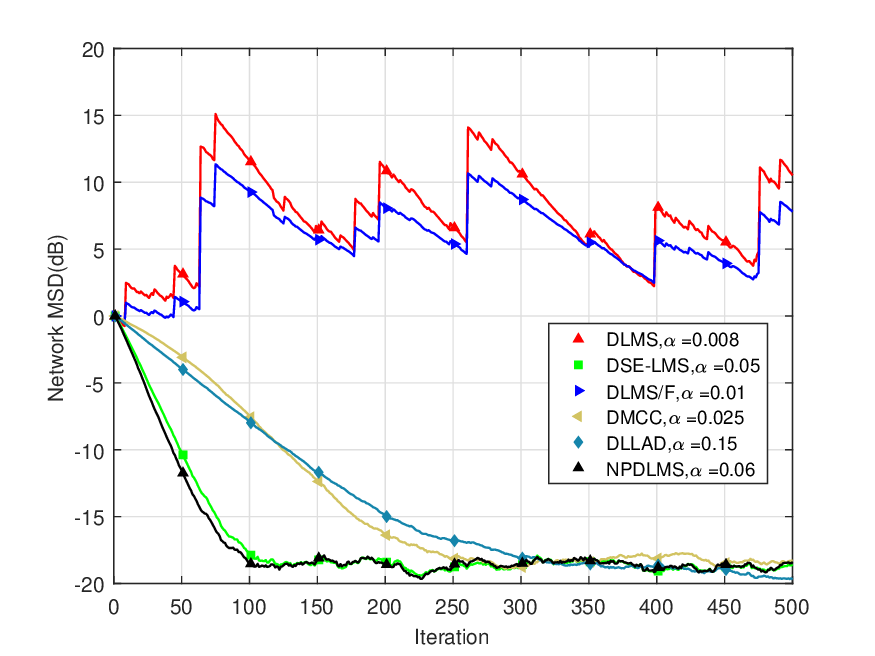}
		\label{fig:stationarynongaussNoiseSimSteady}}
	\subfloat[]{
		\includegraphics[width=2.6in]{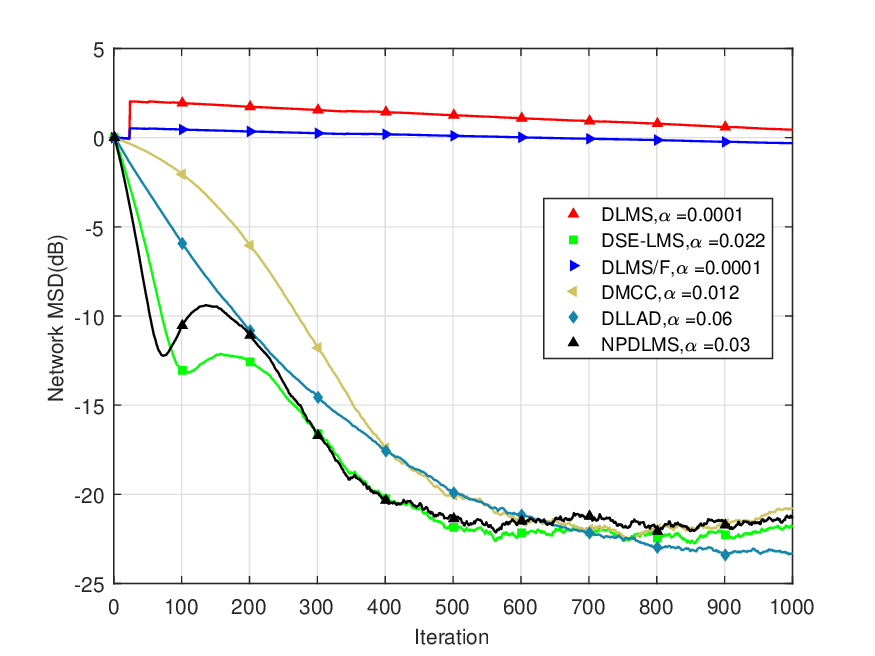}
		\label{fig:nonstationarynongaussNoiseSimSteady}}
	\caption[nonGaussiannoise]{Comparison of algorithms behavior for non-Gaussian noise in similar steady state condition
	\subref{fig:stationarynongaussNoiseSimSteady} stationary environment;
	\subref{fig:nonstationarynongaussNoiseSimSteady} non-stationary environment.}
	\label{fig:nonGaussianNetwork}
\end{figure}

This example examines the performance of algorithms when they have similar steady state MSD. Fig. \ref{fig:stationarynongaussNoiseSimSteady} shows the convergence rate of algorithms in stationary and Fig. \ref{fig:nonstationarynongaussNoiseSimSteady} illustrates the results in the non-stationary environments. The  proposed RDLMS and DSE-LMS algorithms are similar and outperform other algorithms.
\subsection{The role of different parameters}
\begin{figure}[h]
	\centering
	\subfloat[]{
		\label{fig:gaus_S}
		\includegraphics[scale=.45]{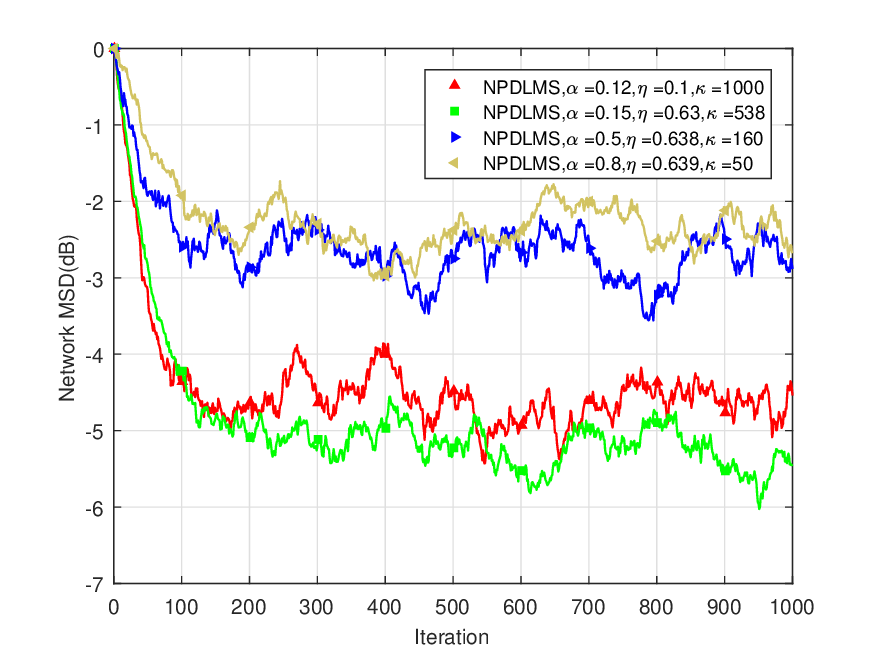}}
	\subfloat[]{
		\label{fig:nonGaus_S}
		\includegraphics[scale=.45]{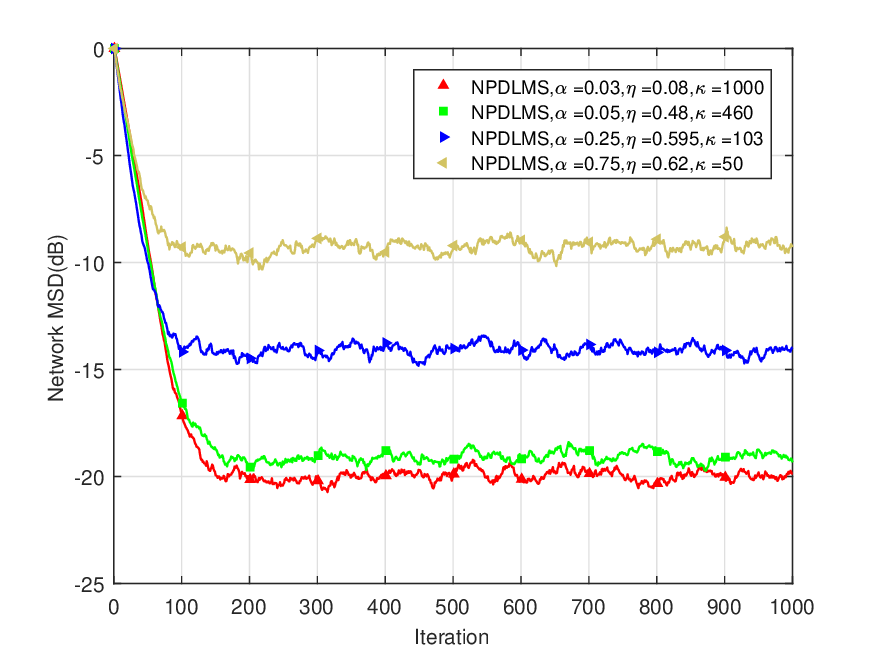}}
	\caption[stationaryNetwork]{Effect of the threshold $\eta$ in stationary environment contaminated by \subref{fig:gaus_S} Gaussian noise with SNR=$-20$ dB, and \subref{fig:nonGaus_S} non-Gaussian noise, $\kappa$ shows the average number of performing the update rule \eqref{eqUpdateRule1}.}
	\label{fig_thrSta}
\end{figure}

\begin{figure}[!h]
	\centering
	\subfloat[]{
		\label{fig:gaus}
		\includegraphics[scale=.45]{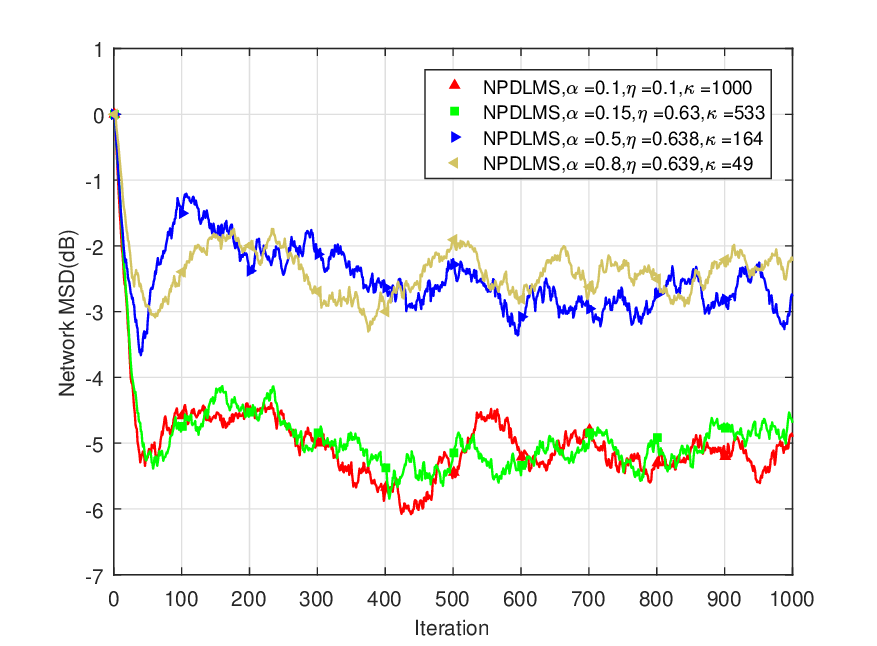}}
	\subfloat[]{
		\label{fig:nonGaus}
		\includegraphics[scale=.45]{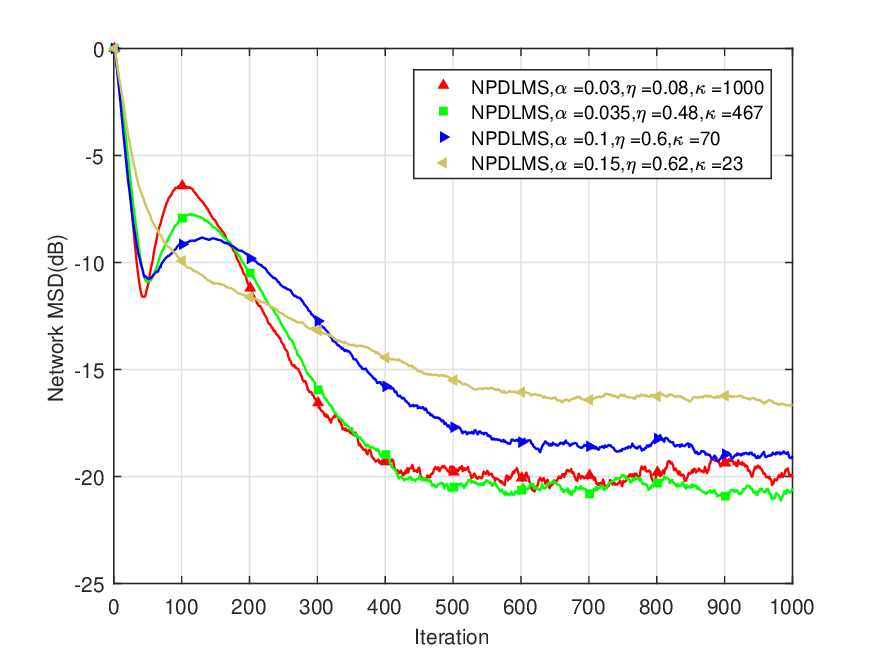}}
	\caption[stationaryNetwork]{Effect of the threshold $\eta$ in non-stationary environment contaminated by \subref{fig:gaus} Gaussian noise with SNR=$-20$ dB, and \subref{fig:nonGaus} non-Gaussian noise, $\kappa$ shows the average number of performing the update rule \eqref{eqUpdateRule1}.}
	\label{fig_thrNonSta}
\end{figure}

First, the role of $\eta$ in \eqref{eqUpdateRulIf} is investigated for Gaussian noise with SNR=$-20$ dB and non-Gaussian noise  $V_{\alpha-stable}(1.2,0,1,0)$ in the stationary (Fig. \ref{fig_thrSta}), and non-stationary (Fig. \ref{fig_thrNonSta}) environments. As seen in two figures, increasing the value of  $\eta$ reduces the number of performing the update rule \eqref{eqUpdateRulIf} and increases the steady-state MSD. However, this performance reduction occurs when reducing the update rule performing less than 100 iterations. In the other words, reducing the performing of \eqref{eqUpdateRulIf} even to half, cannot affect the performance of the algorithm MSD but also increase its execution speed. 

\begin{figure}[h]
	\centering
	\subfloat[]{
		\label{fig:h}
		\includegraphics[scale=.45]{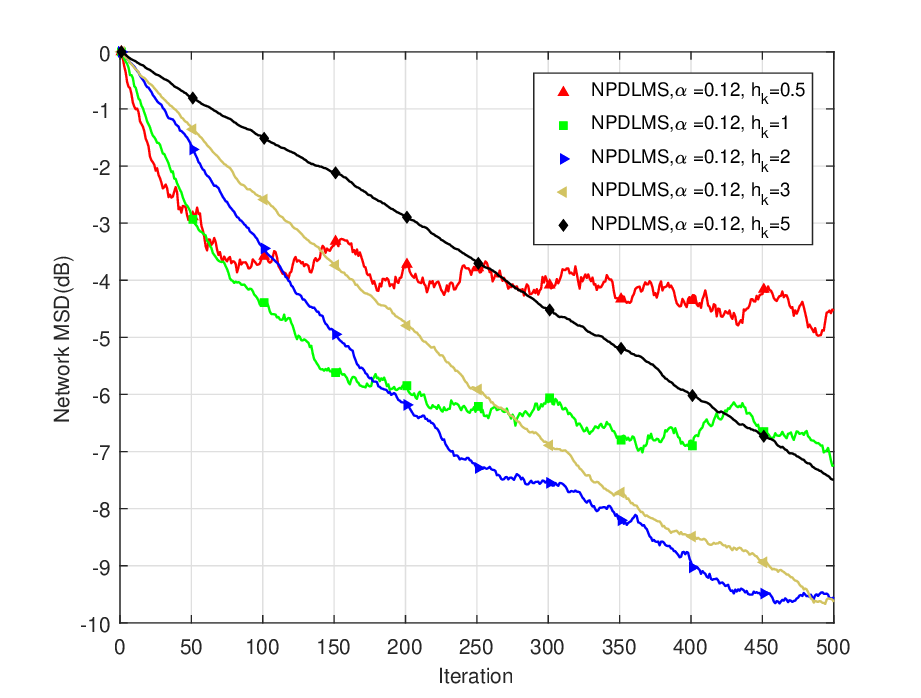}}
	\subfloat[]{
		\label{fig:delta}
		\includegraphics[scale=.45]{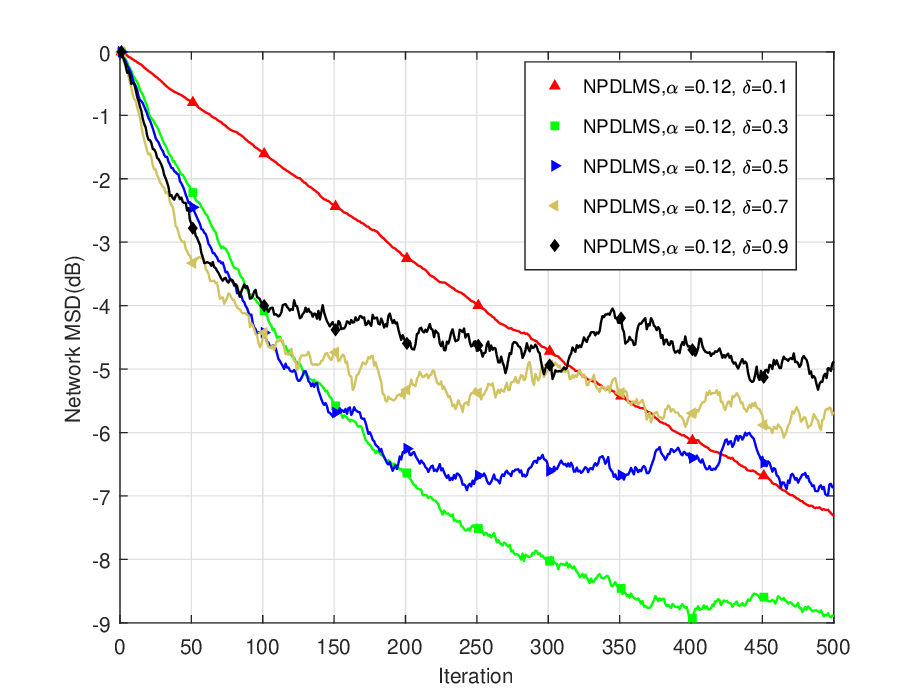}}
	\vspace{-2pt}
	\subfloat[]{
		\label{fig:sigma}
		\includegraphics[scale=.48]{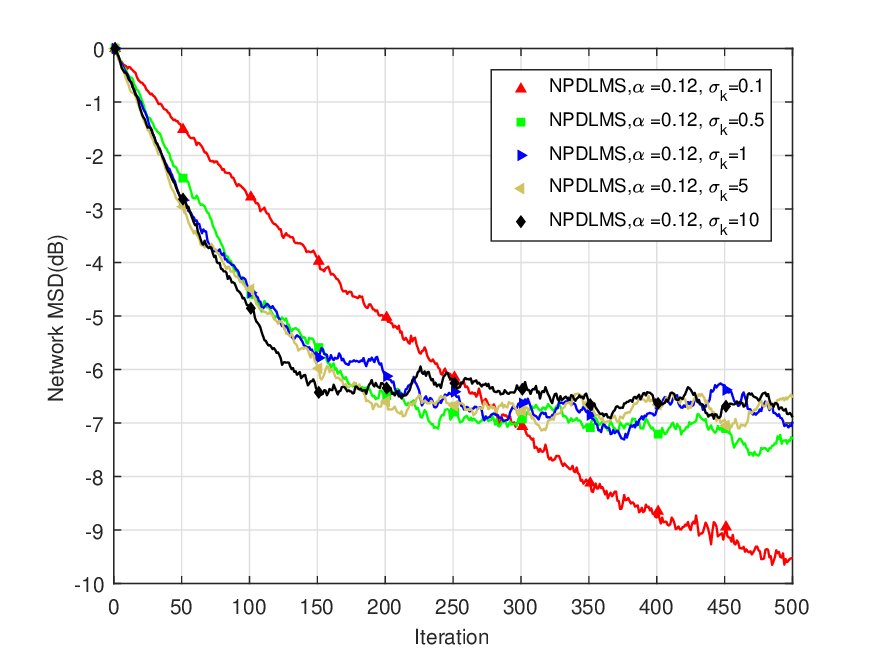}}
	\caption[stationaryNetwork]{Effect of parameters \subref{fig:h} $h$, \subref{fig:delta} $\delta$ and \subref{fig:sigma} $\sigma$ in \eqref{eqGradiantJhat2} in the stationary environment contaminated by Gaussian noise with SNR=-20 d‌B.}
	\label{fig_thr}
\end{figure}

The role of $\delta$, $\sigma$ and $h$ in \eqref{eqGradiantJhat2} is presented in Fig. \ref{fig_thr} for Gaussian noise with SNR=$-20$ dB. 
As seen in Fig. \ref{fig:h}, increasing the value of $h$ improves the steady-state performance but reduces the convergence rate. Changing the value of  $\delta$ shows better performance for smaller values as seen in \ref{fig:delta}.  Also, values grater than 0.1 leads to similar performance of the algorithm according to Fig. \ref{fig:sigma}. 

%
%

\section{ Conclusion}
The present paper proposed to use the maximum a posteriori framework and kernel density estimation to incorporate non-parametric knowledge in designing of the diffusion least mean square adaptive filter. A pseudo-Huber loss function has also been utilized to design the error function and consequently the likelihood distribution. In addition, an error thresholding function has been defined to reduces the computational overhead by removing unnecessary updates.  
Examining the proposed NPDLMS algorithm against Gaussian and non-Gaussian noise indicates its robustness in the stationary and non-stationary environment. Performance analysis of the  NPDLMS algorithm shows the mean and mean-square convergence, stability and transient behavior. Also, the compatibility of the theoretical mean square deviation with the experimental result has been illustrated. 
When the environment is contaminated by non-Gaussian noise whether it is stationary or not, the performance of the NPDLMS is similar to the DSE-LMS and these two algorithms outperform DLMS, DLMS/F, and DMCC.  
As expected, the standard DLMS is the best one in Gaussian noise, while the proposed NPDLMS is similar to it and outperforms other algorithms in high SNR. In addition, the NPDLMS and DSE-LMS convergence outperform other competitors when the steady-state of all algorithms are similar.

\section*{Appendix} \label{Appendix1}
Without loss of generality assume the buffer length in Fig. \ref{fig_computeMu2} is $‌B$, and $\theta_{l}$ is similar with  $\left\{\boldsymbol{\theta}_{l_j}, j\in‌ \mathcal{J}\subset \mathcal{B}_{1}\right\}, |\mathcal{J}|=r_l$ such that $K_{\sigma_l}(\boldsymbol{\theta}_l-\boldsymbol{\theta}_{l_j})\simeq1$, and $i\in \mathcal{B}_{1}$. Therefor, considering \eqref{eqMu_kli} one can write
\begin{equation}\label{eqMu_kliNonEq}
\boldsymbol\mu_{kli}=\frac{K_{\sigma_k}(\boldsymbol{\theta}_k-\boldsymbol{\theta}_{k_i})}{\sum_{j\in \mathcal{B}_{1}}K_{\sigma_k}(\boldsymbol{\theta}_k-\boldsymbol{\theta}_{k_j})}\le \frac{1}{r_l}
\end{equation}
Also, considering \eqref{eqMu_ki}
\begin{equation}\label{eqMu_kiNonEq}
\boldsymbol\mu_{ki}=\frac{K_{\sigma_k}(\boldsymbol{\theta}_k-\boldsymbol{\theta}_{k_i})}{\sum_{j=1}^{B}K_{\sigma_k}(\boldsymbol{\theta}_k-\boldsymbol{\theta}_{k_j})}\le \frac{1}{B}
\end{equation}

\section*{References}
 \bibliographystyle{elsarticle-num} 
 \bibliography{myRef}

\begin{thebibliography}{10}
\expandafter\ifx\csname url\endcsname\relax
  \def\url#1{\texttt{#1}}\fi
\expandafter\ifx\csname urlprefix\endcsname\relax\def\urlprefix{URL }\fi
\expandafter\ifx\csname href\endcsname\relax
  \def\href#1#2{#2} \def\path#1{#1}\fi

\bibitem{R1}
R.~Caballero-{\'A}guila, A.~Hermoso-Carazo, J.~Linares-P{\'e}rez, Distributed
  fusion filters from uncertain measured outputs in sensor networks with random
  packet losses, Information Fusion 34 (2017) 70--79.

\bibitem{di2011}
P.~Di~Lorenzo, S.~Barbarossa, A.~H. Sayed, Bio-inspired swarming for dynamic
  radio access based on diffusion adaptation, in: Signal Processing Conference,
  2011 19th European, IEEE, 2011, pp. 402--406.

\bibitem{tu2011}
S.-Y. Tu, A.~H. Sayed, Mobile adaptive networks, IEEE Journal of Selected
  Topics in Signal Processing 5~(4) (2011) 649--664.

\bibitem{bazzi2015}
W.~M. Bazzi, A.~L. Pak, A.~Rastegarnia, A.~Khalili, Z.~Yang, Formulation and
  steady-state analysis of diffusion mobile adaptive networks with noisy links,
  IET Signal Processing 9~(9) (2015) 631--637.

\bibitem{cao2010}
X.~Cao, J.~Chen, Y.~Xiao, Y.~Sun, Building-environment control with wireless
  sensor and actuator networks: Centralized versus distributed, IEEE
  Transactions on Industrial Electronics 57~(11) (2010) 3596--3605.

\bibitem{duan2017}
X.~Duan, C.~Zhao, S.~He, P.~Cheng, J.~Zhang, Distributed algorithms to compute
  walrasian equilibrium in mobile crowdsensing, IEEE Transactions on Industrial
  Electronics 64~(5) (2017) 4048--4057.

\bibitem{chen2010}
J.~Chen, X.~Cao, P.~Cheng, Y.~Xiao, Y.~Sun, Distributed collaborative control
  for industrial automation with wireless sensor and actuator networks, IEEE
  Transactions on Industrial Electronics 57~(12) (2010) 4219--4230.

\bibitem{bai2018}
X.~Bai, Z.~Wang, L.~Zou, F.~E. Alsaadi, Collaborative fusion estimation over
  wireless sensor networks for monitoring co2 concentration in a greenhouse,
  Information Fusion 42 (2018) 119--126.

\bibitem{tu2014}
S.-Y. Tu, A.~H. Sayed, Distributed decision-making over adaptive networks.,
  IEEE Trans. Signal Processing 62~(5) (2014) 1054--1069.

\bibitem{khawatmi2017}
S.~Khawatmi, X.~Huang, A.~M. Zoubir, Distributed decision-making over mobile
  adaptive networks, in: Acoustics, Speech and Signal Processing (ICASSP), 2017
  IEEE International Conference on, IEEE, 2017, pp. 3864--3868.

\bibitem{cattivelli2011}
F.~S. Cattivelli, A.~H. Sayed, Modeling bird flight formations using diffusion
  adaptation, IEEE Transactions on Signal Processing 59~(5) (2011) 2038--2051.

\bibitem{R8}
C.~G. Lopes, A.~H. Sayed, Diffusion least-mean squares over adaptive networks:
  Formulation and performance analysis, IEEE Transactions on Signal Processing
  56~(7) (2008) 3122--3136.

\bibitem{R9}
F.~S. Cattivelli, C.~G. Lopes, A.~H. Sayed, Diffusion recursive least-squares
  for distributed estimation over adaptive networks, IEEE Transactions on
  Signal Processing 56~(5) (2008) 1865--1877.

\bibitem{R21}
F.~S. Cattivelli, A.~H. Sayed, Diffusion lms strategies for distributed
  estimation, IEEE Transactions on Signal Processing 58~(3) (2010) 1035--1048.

\bibitem{LEE2015}
J.-W. Lee, S.-E. Kim, W.-J. Song, Data-selective diffusion lms for reducing
  communication overhead, Signal Processing 113 (2015) 211 -- 217.
\newblock \href
  {http://dx.doi.org/https://doi.org/10.1016/j.sigpro.2015.01.019}
  {\path{doi:https://doi.org/10.1016/j.sigpro.2015.01.019}}.

\bibitem{chen2014}
J.~Chen, C.~Richard, A.~H. Sayed, Multitask diffusion adaptation over networks,
  IEEE Transactions on Signal Processing 62~(16) (2014) 4129--4144.

\bibitem{chen2015}
J.~Chen, C.~Richard, A.~H. Sayed, Diffusion lms over multitask networks, IEEE
  Transactions on Signal Processing 63~(11) (2015) 2733--2748.

\bibitem{cattivelli2008}
F.~S. Cattivelli, C.~G. Lopes, A.~H. Sayed, Diffusion recursive least-squares
  for distributed estimation over adaptive networks, IEEE Transactions on
  Signal Processing 56~(5) (2008) 1865--1877.

\bibitem{wen2013}
F.~Wen, Diffusion least-mean p-power algorithms for distributed estimation in
  alpha-stable noise environments, Electronics letters 49~(21) (2013)
  1355--1356.

\bibitem{ni2016}
J.~Ni, J.~Yang, Variable step-size diffusion least mean fourth algorithm for
  distributed estimation, Signal Processing 122 (2016) 93--97.

\bibitem{ni2016b}
J.~Ni, J.~Chen, X.~Chen, Diffusion sign-error lms algorithm: Formulation and
  stochastic behavior analysis, Signal Processing 128 (2016) 142--149.

\bibitem{seo2016diffusion}
J.-H. Seo, S.~M. Jung, P.~Park, A diffusion subband adaptive filtering
  algorithm for distributed estimation using variable step size and new
  combination method based on the msd, Digital Signal Processing 48 (2016)
  361--369.

\bibitem{shi2017two}
L.~Shi, H.~Zhao, Two diffusion proportionate sign subband adaptive filtering
  algorithms, Circuits, Systems, and Signal Processing 36~(10) (2017)
  4242--4259.

\bibitem{gao2018steady}
Y.~Gao, J.~Ni, J.~Chen, X.~Chen, Steady-state and stability analyses of
  diffusion sign-error lms algorithm, Signal Processing 149 (2018) 62--67.

\bibitem{chen2018}
F.~Chen, T.~Shi, S.~Duan, L.~Wang, J.~Wu, Diffusion least logarithmic absolute
  difference algorithm for distributed estimation, Signal Processing 142 (2018)
  423--430.

\bibitem{bazzi2015robust}
W.~M. Bazzi, A.~Rastegarnia, A.~Khalili, A robust diffusion adaptive network
  based on the maximum correntropy criterion, in: Computer Communication and
  Networks (ICCCN), 2015 24th International Conference on, IEEE, 2015, pp.
  1--4.

\bibitem{ma2016}
W.~Ma, B.~Chen, J.~Duan, H.~Zhao, Diffusion maximum correntropy criterion
  algorithms for robust distributed estimation, Digital Signal Processing 58
  (2016) 10--19.

\bibitem{li2013}
C.~Li, P.~Shen, Y.~Liu, Z.~Zhang, Diffusion information theoretic learning for
  distributed estimation over network, IEEE Transactions on Signal Processing
  61~(16) (2013) 4011--4024.

\bibitem{Ilin}
A.~Ilin, T.~Raiko, Practical approaches to principal component analysis in the
  presence of missing values, Journal of Machine Learning Research 11~(Jul)
  (2010) 1957--2000.

\bibitem{Babadi}
B.~Babadi, N.~Kalouptsidis, V.~Tarokh, Sparls: The sparse rls algorithm, IEEE
  Transactions on Signal Processing 58~(8) (2010) 4013--4025.
\newblock \href {http://dx.doi.org/10.1109/TSP.2010.2048103}
  {\path{doi:10.1109/TSP.2010.2048103}}.

\bibitem{Arenas}
J.~Arenas-Garcia, A.~R. Figueiras-Vidal, A.~H. Sayed, Mean-square performance
  of a convex combination of two adaptive filters, IEEE Transactions on Signal
  Processing 54~(3) (2006) 1078--1090.
\newblock \href {http://dx.doi.org/10.1109/TSP.2005.863126}
  {\path{doi:10.1109/TSP.2005.863126}}.

\bibitem{Candy}
J.~V. Candy, Bayesian signal processing: classical, modern, and particle
  filtering methods, Vol.~54, John Wiley and Sons, 2016.

\bibitem{R17}
Z.~Chen, S.~L. Gay, S.~Haykin, Proportionate adaptation: New paradigms in
  adaptive filters, Least-mean-square adaptive filters (2003) 293--334.

\bibitem{R18}
K.~T. Andersen, T.~van Waterschoot, M.~Moonen, Adaptive linear prediction
  filters based on maximum a posteriori estimation, in: Signal Processing
  Conference (EUSIPCO), 2015 23rd European, IEEE, 2015, pp. 2706--2710.

\bibitem{R19}
S.~D. Georgiadis, P.~O. Ranta-aho, M.~P. Tarvainen, P.~A. Karjalainen,
  Single-trial dynamical estimation of event-related potentials: a kalman
  filter-based approach, IEEE Transactions on Biomedical Engineering 52~(8)
  (2005) 1397--1406.

\bibitem{guan2019}
S.~Guan, C.~Meng, B.~Biswal, Diffusion probabilistic lms algorithm, arXiv
  preprint arXiv:1908.09730.

\bibitem{ashkezari2018}
S.~Ashkezari-Toussi, H.~Sadoghi-Yazdi, Incorporating nonparametric knowledge to
  the least mean square adaptive filter, Circuits, Systems, and Signal
  Processing 38~(5) (2019) 2114--2137.

\bibitem{hartley2003}
R.~Hartley, A.~Zisserman, Multiple view geometry in computer vision, Cambridge
  university press, 2003.

\bibitem{ashkezari2019}
S.~Ashkezari-Toussi, H.~Sadoghi-Yazdi, Robust diffusion lms over adaptive
  networks, Signal Processing 158 (2019) 201--209.

\bibitem{R2}
A.~H. Sayed, Adaptive networks, Proceedings of the IEEE 102~(4) (2014)
  460--497.

\bibitem{R23}
B.~Widrow, S.~D. Stearns, J.~C. Burgess, Adaptive signal processing edited by
  bernard widrow and samuel d. stearns, Acoustical Society of America Journal
  80 (1986) 991--992.

\bibitem{R24}
N.~Takahashi, I.~Yamada, A.~H. Sayed, Diffusion least-mean squares with
  adaptive combiners: Formulation and performance analysis, IEEE Transactions
  on Signal Processing 58~(9) (2010) 4795--4810.

\bibitem{zheng2017}
Z.~Zheng, Z.~Liu, M.~Huang, Diffusion least mean square/fourth algorithm for
  distributed estimation, Signal Processing 134 (2017) 268--274.

\bibitem{shao1993}
M.~Shao, C.~L. Nikias, Signal processing with fractional lower order moments:
  stable processes and their applications, Proceedings of the IEEE 81~(7)
  (1993) 986--1010.

\bibitem{R25}
A.~H. Sayed, Adaptive filters, John Wiley \& Sons, 2011.

\end{thebibliography}


\end{document}